\documentclass[10pt,twocolumn,letterpaper]{article}

\usepackage[pagenumbers]{cvpr} %

\usepackage{graphicx}
\usepackage{amsmath}
\usepackage{amssymb}
\usepackage{booktabs}

\usepackage{silence}
\usepackage[pagebackref,breaklinks,colorlinks]{hyperref}
\usepackage[capitalize]{cleveref}
\crefname{section}{Sec.}{Secs.}
\Crefname{section}{Section}{Sections}
\Crefname{table}{Table}{Tables}
\crefname{table}{Tab.}{Tabs.}

\def\cvprPaperID{6966} %
\def\confName{CVPR}
\def\confYear{2023}

\makeatletter
\apptocmd\@maketitle{{\teaserfigure{}\par}}{}{}
\makeatother

\newcommand{\imagewidth}{2.3cm}
\newcommand{\imageheight}{3.5263671874999996cm}
\newcommand{\smallimagewidth}{0.8cm}
\newcommand{\smallimageheight}{1.2265625cm}
\newcommand{\borderthickness}{0.41pt}
\def\myshift#1{\raisebox{0.5ex}}
\newcommand{\teasernobox}{
\begin{subfigure}[b]{\linewidth}
    \centering
    \resizebox{\linewidth}{!}{
    \begin{tikzpicture}[
        >=stealth',
        overlay/.style={
          anchor=south west, 
          draw=black,
          rectangle, 
          line width=0.8pt,
          outer sep=0,
          inner sep=0,
        },
    ]
        \small
        \matrix[matrix of nodes, column sep=1pt, row sep=1pt, ampersand replacement=\&, inner sep=0, outer sep=0] (training) {
            \includegraphics[height=\imageheight]{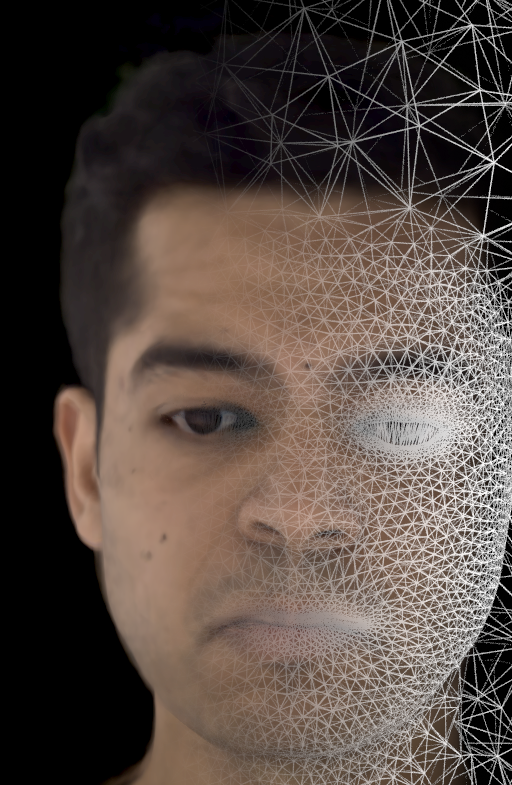} \&
            \includegraphics[height=\imageheight]{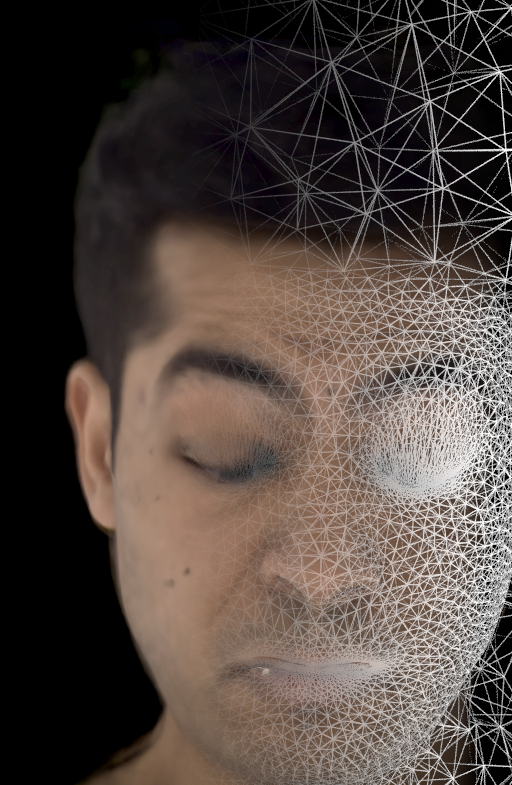} \&
            \includegraphics[height=\imageheight]{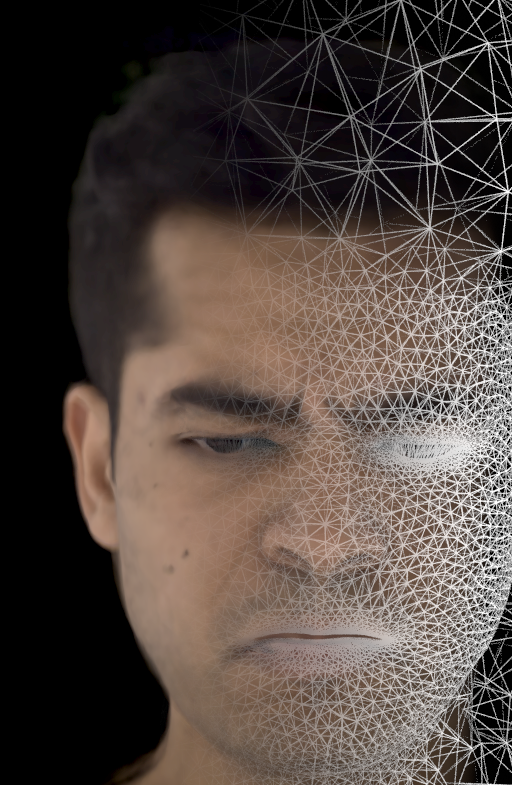} \&
            \includegraphics[height=\imageheight]{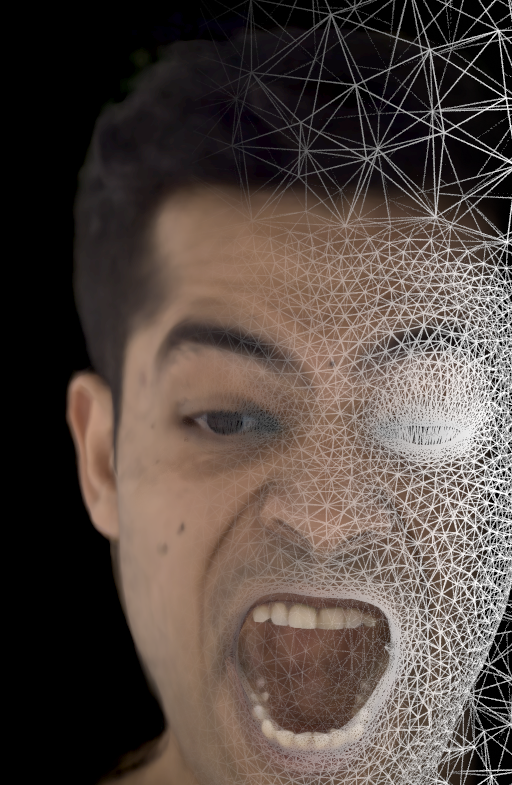} \&
            \includegraphics[height=\imageheight]{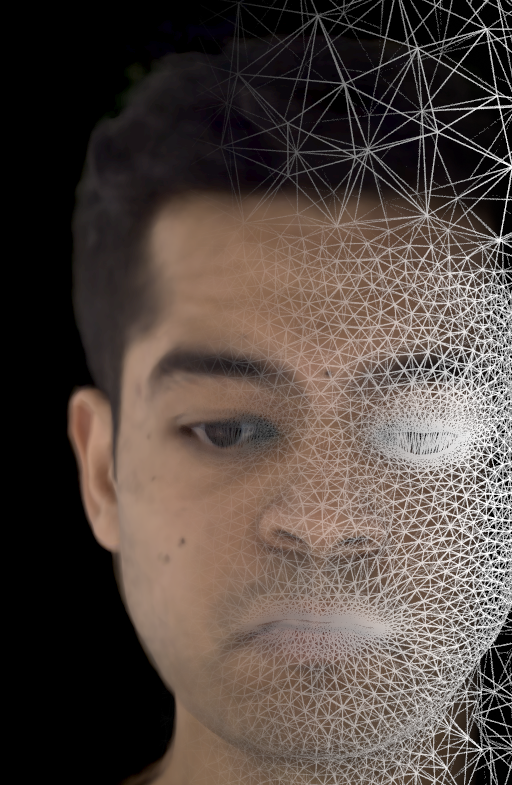} \\
            \includegraphics[height=\imageheight]{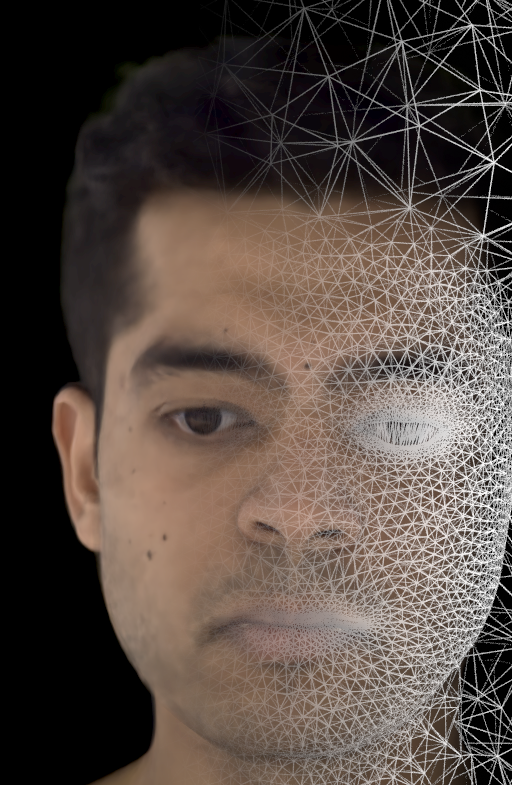} \&
            \includegraphics[height=\imageheight]{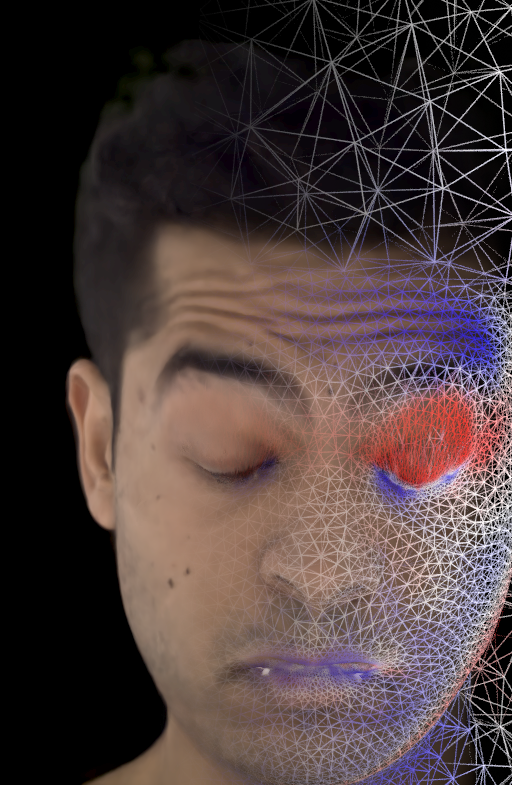} \&
            \includegraphics[height=\imageheight]{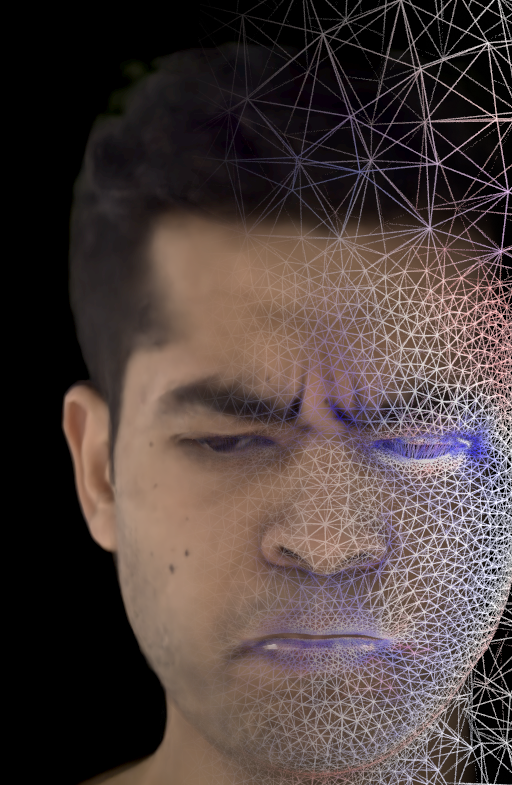} \&
            \includegraphics[height=\imageheight]{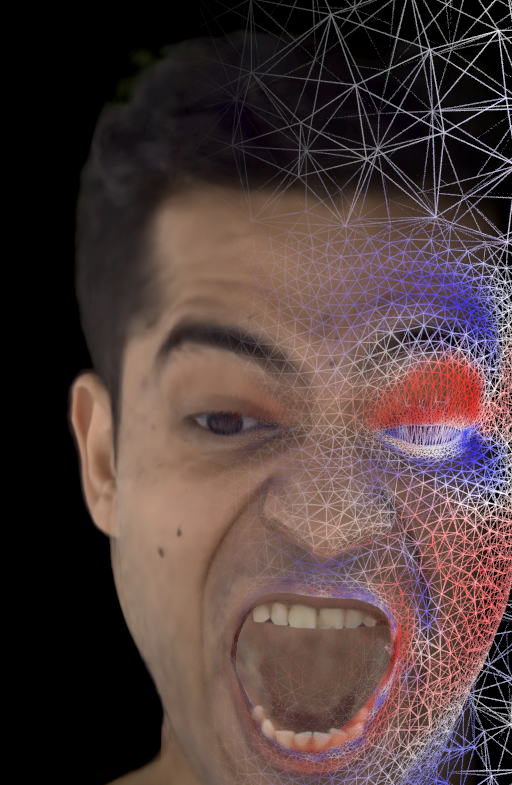} \&
            \includegraphics[height=\imageheight]{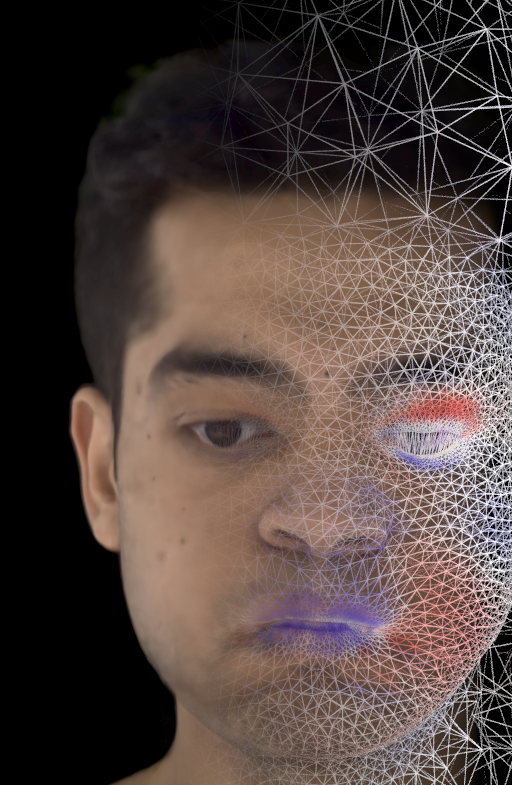} \\
        };
        \node[above right=\borderthickness and \borderthickness of training-2-1.south west, overlay]{
            \includegraphics[height=\smallimageheight]{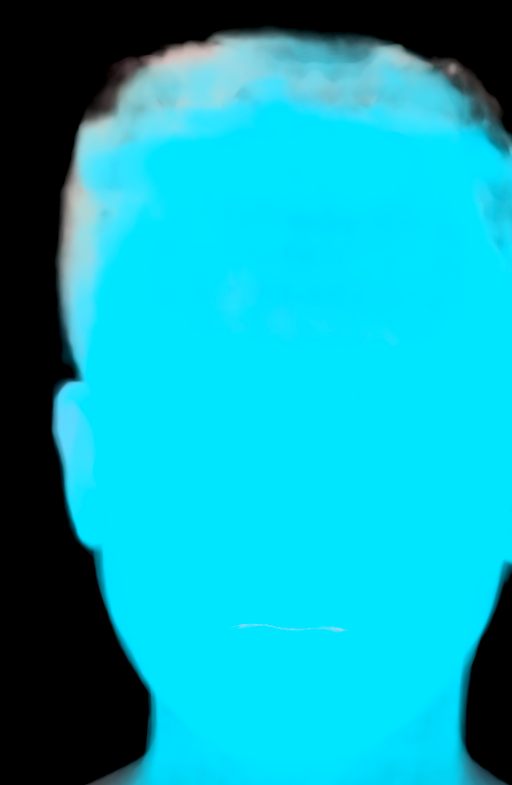}
        };
        \node[above right=\borderthickness and \borderthickness of training-2-2.south west, overlay]{
            \includegraphics[height=\smallimageheight]{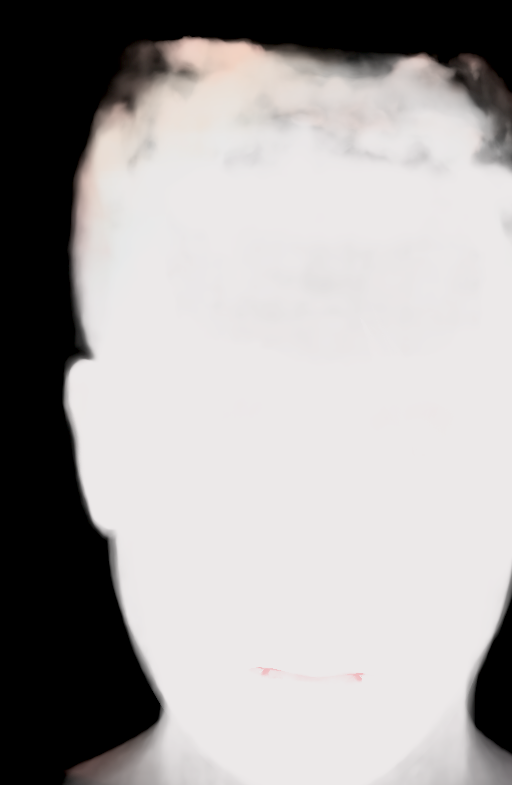}
        };
        \node[above right=\borderthickness and \borderthickness of training-2-3.south west, overlay]{
            \includegraphics[height=\smallimageheight]{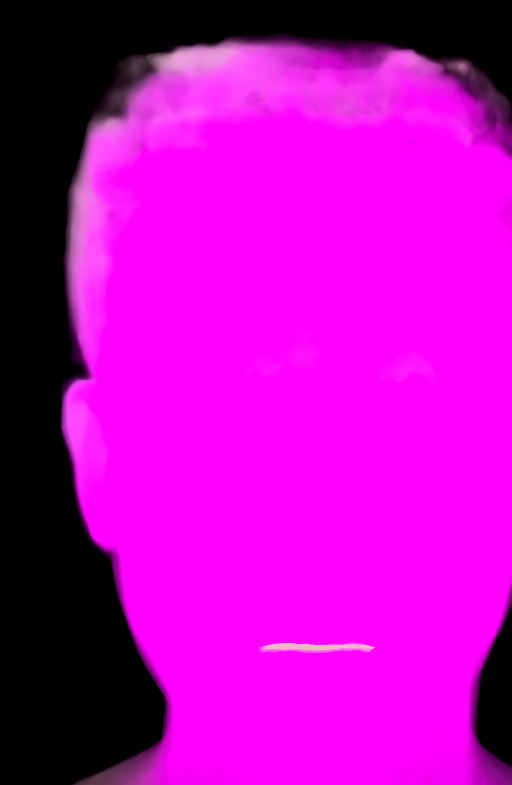}
        };
        \node[above right=\borderthickness and \borderthickness of training-2-4.south west, overlay]{
            \includegraphics[height=\smallimageheight]{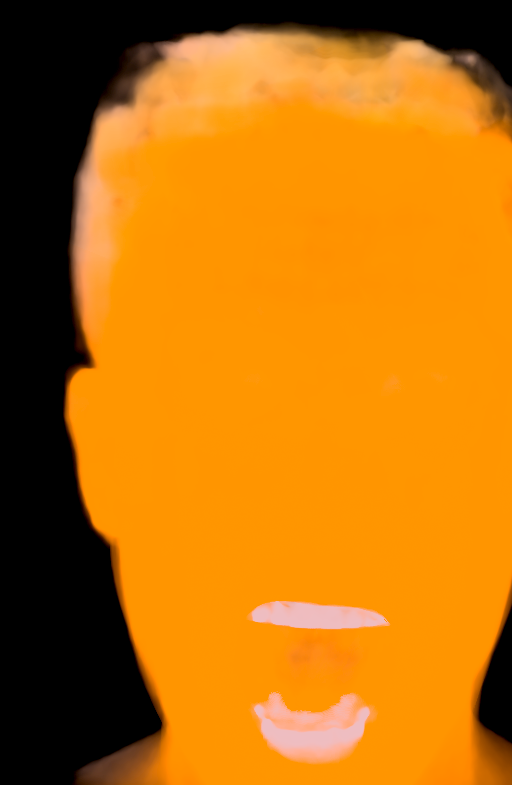}
        };
        \node[above right=\borderthickness and \borderthickness of training-2-5.south west, overlay]{
            \includegraphics[height=\smallimageheight]{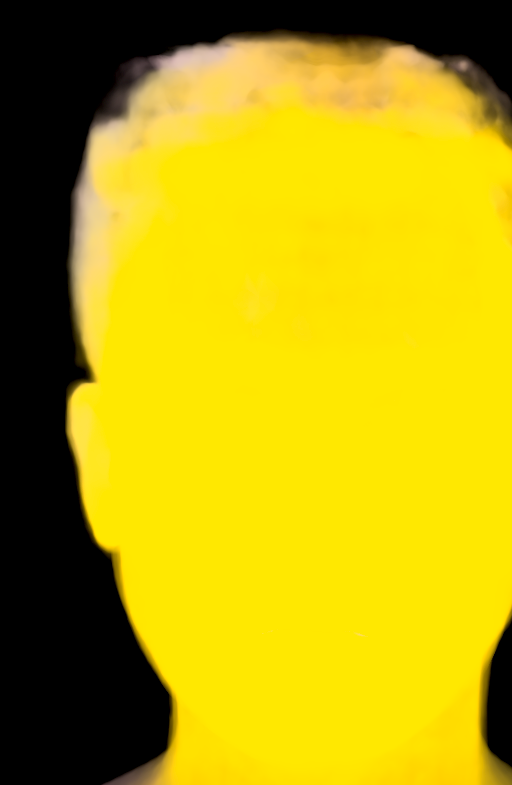}
        };
        
        \matrix[matrix of nodes, column sep=1pt, row sep=1pt, ampersand replacement=\&, right=1em of training, inner sep=0, outer sep=0] (final) {
            \includegraphics[height=\imageheight]{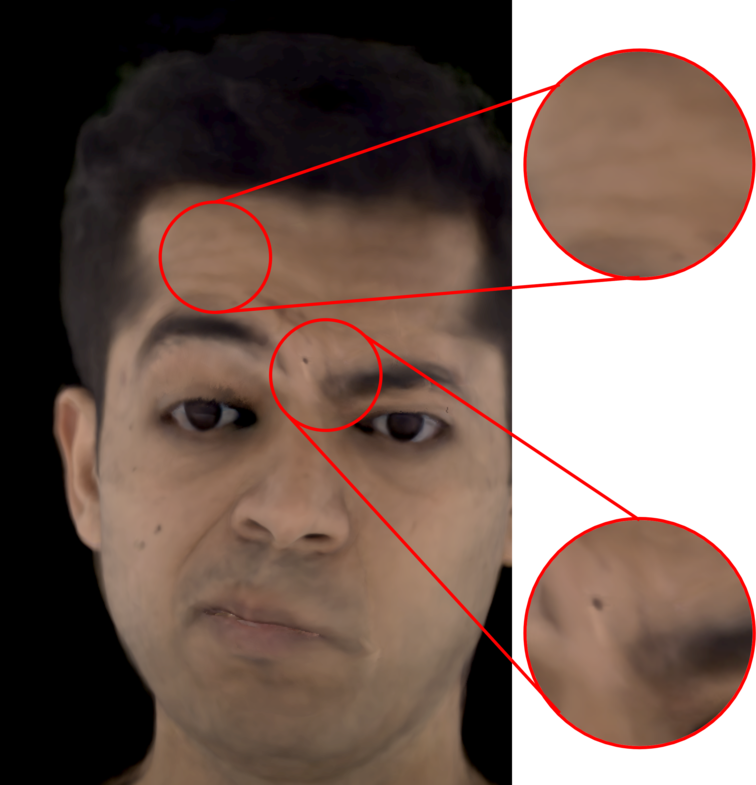}\\
            \includegraphics[height=\imageheight]{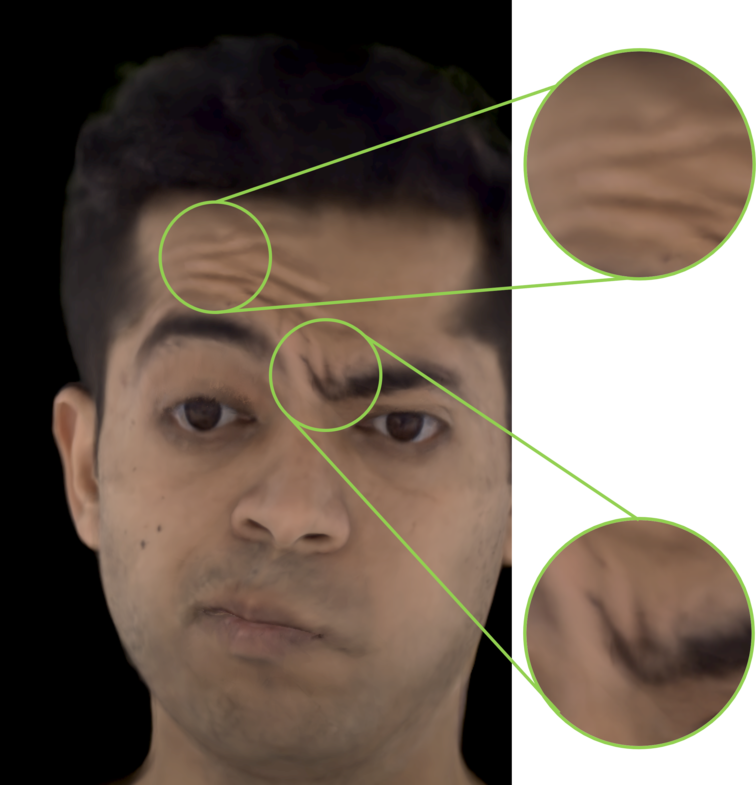}  \\ 
        };
        \node[above right=\borderthickness and \borderthickness of final-2-1.south west, overlay]{
            \includegraphics[height=\smallimageheight]{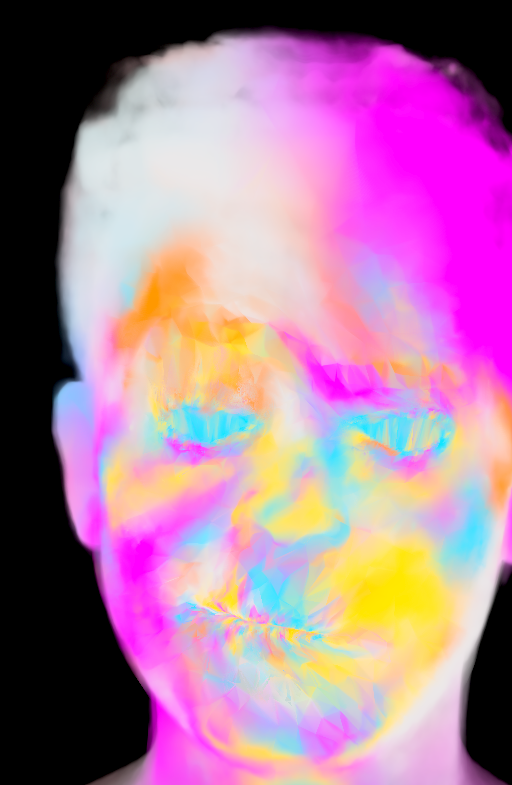}
        };
        \node[above right=\borderthickness and \borderthickness of final-1-1.south west, overlay]{
            \includegraphics[height=\smallimageheight]{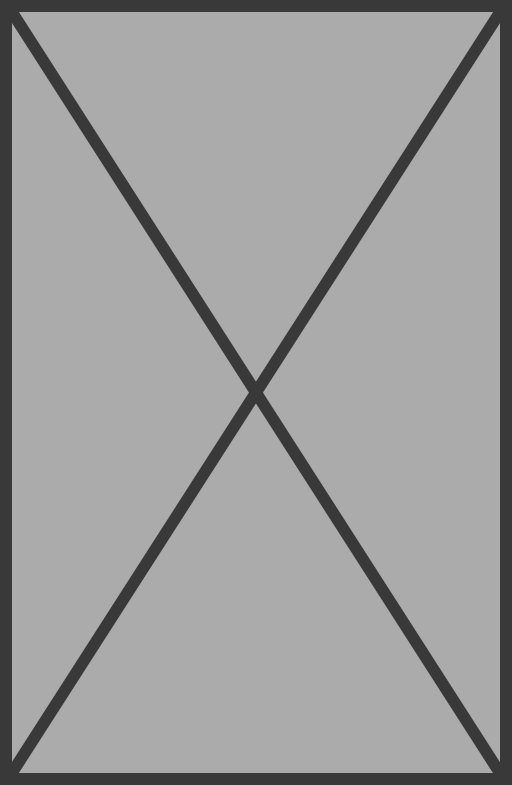}
        };
        
        \node[left=0.75em of training-1-1.west, rotate=90, anchor=center] {VolTeMorph};
        \node[left=0.75em of training-2-1.west, rotate=90, anchor=center] {\textbf{Ours}};
        \node[below=0.25em of training.south] {Rendered training expressions};
        \node[below left=0.25em and -2.5em of final-2-1.south] {Novel expression};
        
        \draw[densely dotted, line width=0.1em] ($(training-1-5.north east)+(0.5em,0.0em)$) -- ($(training-2-5.south east)+(0.5em,0.0em)$);

    \end{tikzpicture}
    }
\end{subfigure}
}

\newcommand{\teaserfigure}{
\vspace{-5mm}
\captionsetup{type=figure}

\teasernobox

\vspace{-2mm}
\setcounter{figure}{0} %
\captionsetup{type=figure}
\captionof{figure}{
\textbf{Teaser} --
Given five multi-view frames of different expressions, our approach generates a model capable of capturing the fine-grained details of a novel expression beyond the resolution of the underlying face model~\cite{garbin2022voltemorph} (top right corner).
This is achieved by \textit{blending} the radiance fields computed for individual expressions, where the blending coefficients are modulated accordingly to \textit{local} volumetric changes. 
These volumetric changes are measured as the difference in the tetrahedral volume of a mesh that deforms with the expression (\mycoloredbox{seismicred} increase, \mycoloredbox{seismicblue} decrease, and \mycoloredbox{seismicgray} no change in volume).
Such an approach allows \textit{\methodname{}} to render sharp, expression-dependent details of the face without increasing the resolution of the mesh (bottom right corner). %
}
\label{fig:teaser}
\vspace{1em}
}

\makeatletter
\@namedef{ver@everyshi.sty}{}
\makeatother

\usepackage{float}
\usepackage{bm}

\usepackage{booktabs}
\usepackage{multirow}
\usepackage{caption}
\usepackage{subcaption}
\usepackage[percent]{overpic}
\usepackage{amssymb}%
\usepackage{pifont}%
\usepackage{cleveref}
\usepackage{colortbl}
\usepackage[usenames,dvipsnames]{xcolor}
\usepackage[accsupp]{axessibility}

\usepackage{tikz}
\usepackage{amsmath}
\usepackage{bbm}
\usepackage{silence}
\usetikzlibrary{matrix, positioning,arrows,arrows.meta,calc,decorations.pathreplacing,decorations.text,spy}

\tikzset{
  img/.style={
    inner sep=0pt,     %
    outer sep=0pt,     %
    rectangle,
    align=center} %
}

\definecolor{turquoise}{cmyk}{0.65,0,0.1,0.3}
\definecolor{purple}{rgb}{0.65,0,0.65}
\definecolor{dark_green}{rgb}{0, 0.5, 0}
\definecolor{orange}{rgb}{0.8, 0.6, 0.2}
\definecolor{red}{rgb}{0.8, 0.2, 0.2}
\definecolor{darkred}{rgb}{0.6, 0.1, 0.05}
\definecolor{blueish}{rgb}{0.0, 0.3, .6}
\definecolor{light_gray}{rgb}{0.7, 0.7, .7}
\definecolor{pink}{rgb}{0.8, 0, 0.8}
\definecolor{greyblue}{rgb}{0.25, 0.25, 1}

\definecolor{thirdbestcolor}{rgb}{1,1, 0.6}
\definecolor{secondbestcolor}{rgb}{1, 0.9, 0.6}
\definecolor{firstbestcolor}{rgb}{1, 0.6, 0.6}

\definecolor{seismicblue}{rgb}{0.0,0.0,0.75}
\definecolor{seismicred}{rgb}{0.62,0.03,0.0}
\definecolor{seismicgray}{rgb}{0.9,0.9,0.9}

\newcommand\mycoloredbox[1]{\textcolor{#1}{\rule{0.5em}{0.5em}}}
\newcommand{\cmark}{\textcolor{green}{\ding{51}}}%
\newcommand{\xmark}{\textcolor{red}{\ding{55}}}%
\newcommand{\supplementary}{\texttt{Supplementary}}

\newcommand{\expect}{\mathbb{E}}
\newcommand{\real}{\mathbb{R}}

\usepackage{enumitem}
\setlist[itemize]{nosep,leftmargin=.15in,topsep=0em}
\setlist[enumerate]{nosep,leftmargin=*,topsep=0em}

\renewcommand{\paragraph}[1]{\vspace{.5em}\noindent\textbf{#1}.}

\newcommand{\eq}[1]{(\ref{eq:#1})}

\newcommand{\VolTeMorph}{VolTeMorph~\cite{garbin2022voltemorph}\xspace}
\newcommand{\methodname}{BlendFields}
\newcommand{\iExpr}{k}
\newcommand{\nExpr}{K}
\newcommand{\expression}{\mathbf{e}}
\newcommand{\pixelcolor}{C}
\newcommand{\blendfield}{\boldsymbol{\alpha}}
\newcommand{\blendingweight}{\alpha}
\newcommand{\vertex}{\mathbf{v}}
\newcommand{\indicator}{\mathbbm{1}}
\newcommand{\temp}{\tau}
\newcommand{\pos}{\mathbf{x}}

\newcommand{\map}{\mathcal{T}}

\newcommand{\density}{\ensuremath{\sigma}}
\newcommand{\radiance}{\mathbf{c}}
\newcommand{\ray}{\ensuremath{\mathbf{r}}}
\newcommand{\template}[1]{\bar{#1}}
\newcommand{\aux}[1]{\tilde{#1}}
\newcommand{\loss}{\mathcal{L}}

\newcommand{\image}{C}
\newcommand{\tet}{\mathbf{T}}
\newcommand{\edgematrix}{\mathbf{D}}

\newcommand{\meshVertices}{\mathbf{V}}

\newcommand{\geometry}{\mathcal{G}}
\newcommand{\laplacian}{\ensuremath{\mathbf{L}}}
\newcommand{\hyperparam}{\lambda}

\newcommand{\outputcolor}{\ensuremath{\mathbf{c}}}

\newcommand{\viewdirection}{\ensuremath{\mathbf{v}}}

\newcommand{\neighbourhood}{\ensuremath{\mathcal{N}}}

\newcommand{\weightdiffusion}{\ensuremath{\lambda_\text{diff}}}

\newcommand{\numberofsamples}{\ensuremath{N}}
\newcommand{\coarsesamples}{\ensuremath{N_\text{coarse}}}
\newcommand{\importancesamples}{\ensuremath{N_\text{importance}}}

\newcommand{\versionone}{}

\newcommand{\volume}{\mathcal{V}}

\newcommand{\customfootnotetext}[2]{{%
  \renewcommand{\thefootnote}{#1}%
  \footnotetext[0]{#2}}}%

\begin{document}

\title{\methodname: Few-Shot Example-Driven Facial Modeling}

\author{
Kacper Kania$^{1,2,\dagger}$ \quad
Stephan J. Garbin$^{3}$ \quad
Andrea Tagliasacchi$^{4,5,\ddagger}$ \quad
Virginia Estellers$^{3}$ \quad\\
Kwang Moo Yi$^{2}$ \quad
Julien Valentin$^{3}$ \quad
Tomasz Trzci\'{n}ski$^{1,6,7}$ \quad
Marek Kowalski$^{3}$ \\[.5em]
Warsaw University of Technology$^{1}$\quad
University of British Columbia$^{2}$\quad
Microsoft$^{3}$\\
Simon Fraser University$^{4}$\quad
Google Brain$^{5}$\quad
IDEAS NCBR$^{6}$\quad
Tooploox$^{7}$\quad
}

\maketitle
\customfootnotetext{$\dagger$}{Work done during an internship at Microsoft Research Cambridge.}
\customfootnotetext{$\ddagger$}{Work done at Simon Fraser University.}
\begin{abstract}
\vspace{-1em}
Generating faithful visualizations of human faces requires capturing both coarse and fine-level details of the face geometry and appearance. 
Existing methods are either data-driven, requiring an extensive corpus of data not publicly accessible to the research community, or fail to capture fine details because they rely on geometric face models that cannot represent fine-grained details in texture with a mesh discretization and linear deformation designed to model only a coarse face geometry.
We introduce a method that bridges this gap by drawing inspiration from traditional computer graphics techniques. Unseen expressions are modeled by blending appearance from a sparse set of extreme poses.
This blending is performed by measuring local volumetric changes in those expressions and locally reproducing their appearance whenever a similar expression is performed at test time.
We show that our method generalizes to unseen expressions, adding fine-grained effects on top of smooth volumetric deformations of a face, and demonstrate how it generalizes beyond faces.
\end{abstract}
\vspace{-2em}

\section{Introduction}
\label{sec:intro}

\begin{table*}[!t]
\centering
\resizebox{\linewidth}{!}{
    \begin{tabular}{lccccccc|c}
        \toprule
                                            & NeRF~\cite{mildenhall2020nerf}    & NeRFies~\cite{park2021nerfies} & HyperNeRF~\cite{park2021hypernerf}     & NeRFace~\cite{gafni2021dynamic}& NHA~\cite{neural_head_avatars} & AVA \cite{cao2022authentic}  & VolTeMorph~\cite{garbin2022voltemorph}     & \textbf{Ours}\\
        \midrule
        Applicability beyond faces             & \cmark                            & \cmark                         & \cmark                                 & \xmark                         & \xmark                         & \xmark                       & \cmark                                     & \cmark \\
        Interpretable control               & \xmark                            & \xmark                         & \xmark                                 & \cmark                         & \cmark                         & \xmark                       & \cmark                                     & \cmark \\
        Data efficiency                      & \xmark                            & \cmark                         & \cmark                                 & \xmark                         & \cmark                         & \xmark                       & \cmark                                     & \cmark\\
        Expression-dependent changes        & \xmark                            & \xmark                         & \cmark                                 & \cmark                         & \cmark                         & \cmark                       & \xmark                                     & \cmark \\
        Generalizability to unknown expressions  & \xmark                            & \xmark                         & \xmark                                 & \cmark                         & \cmark                         & \xmark                       & \cmark                                     & \cmark \\
        \bottomrule
    \end{tabular}
}
\caption{\textbf{Comparison} -- We compare several methods to our approach. Other methods fall short in data efficiency and applicability. For example, AVA~\cite{cao2022authentic} requires 3.1 million 
training images
while VolTeMorph \cite{garbin2022voltemorph} cannot model 
expression-dependent wrinkles realistically. }
\label{tab:ava-ours-comparison}
\vspace{-1.25em}
\end{table*}

Recent advances in neural rendering of 3D scenes~\cite{tewari2022advances} offer 3D reconstructions of unprecedented quality~\cite{mildenhall2020nerf} with an ever-increasing degree of control ~\cite{kania2022conerf, liu2021editing}.
Human faces are of particular interest to the research community~\cite{athar2022rignerf, gafni2021dynamic, garbin2022voltemorph, gao2022reconstructing} due to their application in creating realistic digital doubles~\cite{ma2021pixel, tewari2022advances, zhang2022avatargen, zhi2022dualspace}.

To render facial expressions not observed during training, current solutions~\cite{athar2022rignerf, gafni2021dynamic, garbin2022voltemorph, gao2022reconstructing} rely on \textit{parametric} face models~\cite{blanz1999morphable}.
These allow radiance fields~\cite{mildenhall2020nerf} to be controlled by facial parameters estimated by off-the-shelf face trackers~\cite{li2017flame}.
However, parametric models primarily capture smooth deformations and lead to digital doubles that lack realism because fine-grained and expression-dependent phenomena like wrinkles are not faithfully reproduced.

Authentic Volumetric Avatars (AVA)~\cite{cao2022authentic} overcomes this issue by learning from a large multi-view dataset of synchronized and calibrated images captured under controlled lighting. Their dataset covers a series of dynamic facial expressions and multiple subjects.
However, this dataset remains unavailable to the public and is expensive to reproduce. Additionally, training models from such a large amount of data requires significant compute resources.
To democratize digital face avatars, more efficient solutions in terms of hardware, data, and compute are necessary.

We address the efficiency concerns by building on the recent works in Neural Radiance Fields~\cite{garbin2022voltemorph,xu2022deforming,yuan2022nerf}. %
In particular, we extend VolTeMorph~\cite{garbin2022voltemorph} to render facial details learned from images of a sparse set of expressions.
To achieve this, we draw inspiration from blend-shape correctives~\cite{lewis2014practice}, which are often used in computer graphics as a data-driven way to correct potential mismatches between a simplified model and the complex phenomena it aims to represent.
In our setting, this mismatch is caused by the low-frequency deformations that the tetrahedral mesh from VolTeMorph~\cite{garbin2022voltemorph}, designed for real-time performance, can capture, and the high-frequency nature of expression \mbox{wrinkles}.

We train multiple radiance fields, one for each of the $\nExpr$ sparse expressions present in the input data, and blend them to correct the low-frequency estimate provided by VolTeMorph~\cite{garbin2022voltemorph}; see \cref{fig:teaser}.
We call our method \textit{\methodname{}} since it resembles the way blend shapes are employed in 3DMMs~\cite{blanz1999morphable}.
To keep $\nExpr$ small (\ie, to maintain a few-shot regime), we perform local blending to exploit the known correlation between wrinkles and changes in local differential properties~\cite{irving2004invertible, raman2022mesh}.
Using the dynamic geometry of~\cite{garbin2022voltemorph}, local changes in differential properties can be easily extracted by analyzing the 
tetrahedral representation 
underlying the 
corrective blendfields of our model.

\paragraph{Contributions}
We outline the main qualitative differences between our approach and related works in \cref{tab:ava-ours-comparison}, and our empirical evaluations confirm these advantages. In summary, we:
\begin{itemize}
\item extend VolTeMorph~\cite{garbin2022voltemorph} to enable modeling of high-frequency information, such as expression wrinkles in a few-shot setting; 
\item introduce correctives~\cite{blanz1999morphable} to neural field representations and activate them according to local deformations~\cite{raman2022mesh};
\item 
make this topic more accessible 
with an alternative to techniques that are data and compute-intensive~\cite{cao2022authentic};
\item show that our model generalizes beyond facial modeling, for example, in the modeling of wrinkles on a deformable object made of rubber.
\end{itemize}
\section{Related Works}
\label{ref:related}
\vspace{-0.25em}
Neural Radiance Fields (NeRF)~\cite{mildenhall2020nerf} is a method for generating 3D content from images taken with commodity cameras. It has prompted many follow-up works \cite{kaizhang2020nerfpp,park2021nerfies,martin2021nerf,barron2021mip,barron2022mip,verbin2022ref,tancik2022block,huang2022hdr,suhail2022light,xiangli2021citynerf,mildenhall2022nerf,rematas2022urban} and a major change in the field for its photorealism. The main limitations of NeRF are its rendering speed, being constrained to static scenes, and lack of ways to control the scene. Rendering speed has been successfully addressed by multiple follow-up works~\cite{hedman2021baking,garbin2021fastnerf,plenoctrees}. Works solving the limitation to static scenes\cite{xian2021space,athar2022rignerf,noguchi2022watch,wang2022fourier,attal2021torf,weng2022humannerf,zhao2022humannerf,jiang2022neuman} and adding explicit control~\cite{kania2022conerf,wang2022clip,kim2022ae,cheng2022cross,yuan2022nerf,sun2022fenerf,yang2022neumesh} have limited resolution or require large amounts of training data because they rely on controllable coarse models of the scene (\eg, 3DMM face model~\cite{blanz1999morphable}) or a conditioning signal~\cite{park2021hypernerf}. Methods built on an explicit model are more accessible because they require less training data but are limited by the model's resolution. Our technique finds a sweet spot between these two regimes by using a limited amount of data to learn details missing in the controlled model and combining them together. Our experiments focus on faces because high-quality data and 3DMM face models are publicly available, which are the key component for creating digital humans.

\subsection{Radiance Fields}
Volumetric representations~\cite{vicini2021nonexponential} have grown in popularity because they can represent complex geometries like hair more accurately than mesh-based ones. Neural Radiance Fields (NeRFs)~\cite{mildenhall2020nerf} model a radiance volume with a coordinate-based MLP learned from posed images. The MLP predicts density~$\density(\pos)$ and color~$\outputcolor(\pos, \viewdirection)$ for each point $\pos$ in the volume and view direction~$\viewdirection$ of a given camera.
To supervise the radiance volume with the input images, each image pixel is associated with a ray $\ray(t)$ cast from the camera center to the pixel, and samples along the ray are accumulated to determine the value of the image pixel $\pixelcolor(\ray)$:
\begin{equation}
\label{eq:rendering-equation}
\pixelcolor(\ray)=\int^{t_f}_{t_n}T(t)\:\density(\ray(t))\:\outputcolor(\ray(t), \viewdirection)dt,
\end{equation}
where $t_n$ and $t_f$ are near and far planes, and
\begin{equation}
    T(t) = \exp\left(-\int^t_{t_n}\density(\ray(s))ds\right)
    ,
\end{equation}
is the transmittance function~\cite{tagliasacchi2022volume}. The weights of the MLP are optimized to minimize the mean squared reconstruction error between the target pixel and the output pixel.%
Several methods have shown that replacing the implicit functions approximated with an MLP for a function discretized on an explicit voxel grid results in a significant rendering and training speed-up~\cite{garbin2021fastnerf, hedman2021baking, plenoctrees, directVoxelOptimisation, neuralSparseVoxelFields}.

\subsection{Animating Radiance Fields}
Several works exist to animate the scene represented as a NeRF. D-NeRF uses an implicit deformation model that maps sample positions back to a canonical space~\cite{pumarola2021d}, but it cannot generalize to unseen deformations. Several works \cite{park2021nerfies, park2021hypernerf, gafni2021dynamic, tretschk2021non} additionally account for changes in the observed scenes with a per-image latent code to model changes in color as well as shape, but it is unclear how to generalize the latents when animating a sequence without input images. Similarly, works focusing on faces \cite{gafni2021dynamic,athar2022rignerf,zhuang2021mofanerf,gao2022reconstructing} use parameters of a face model to condition NeRF's MLP, or learn a latent space of images and geometry~\cite{cao2022authentic,wang2022morf,lombardi2019neural,mihajlovic2022keypointnerf,ma2021pixel,lombardi2021mixture} that does not extrapolate beyond expressions seen during training. 

In contrast to these approaches, we focus on using as little temporal training data as possible (\ie five frames) while ensuring generalization. For this reason, we build our method on top of VolTeMorph~\cite{garbin2022voltemorph}, that uses a parametric model of the face to track the deformation of points in a volume around the face and builds a radiance field controlled by the parameters of a 3DMM. After training, the user can render an image for any  expression of the face model. However, the approach cannot generate expression-dependent high-frequency details; see~\cref{fig:teaser}.

Similarly, NeRF-Editing~\cite{yuan2022nerf} and NeRF Cages~\cite{xu2022deforming} propose to use tetrahedral meshes to deform a single-frame NeRF reconstruction.
The resolution of the rendered scenes in these methods is limited by the resolution of the tetrahedral cage, which is constrained to a few thousand elements. 

We discuss additional concurrent works in \supplementary{}.

\subsection{Tetrahedral Cages}
To apply parametric mesh models, it is necessary to extend them to the volume to support the volumetric representation of NeRF. Tetrahedral cages are a common choice for their simplicity and ubiquity in computer graphics~\cite{garbin2022voltemorph,yang2022neumesh,xu2022deforming}. For example, VolTeMorph uses dense landmarks~\cite{wood2022dense} to fit a parametric face model whose blendshapes have been extended to a tetrahedral cage with finite elements method~\cite{clough1960thefe}. 
These cages can be quickly deformed and raytraced~\cite{molino2003tetrahedral} using parallel computation on GPUs~\cite{cook2012cuda} while driving the volume into the target pose and allowing early ray termination for fast rendering. We further leverage the tetrahedral cage and use its differential properties~\cite{irving2004invertible}, such as a local volume change, to model high-frequency details. For example, a change from one expression to another changes the volume of tetrahedra in regions where wrinkle formation takes place while it remains unchanged in flat areas. We can use this change in volume to select which of the trained NeRF expressions should be used for each tetrahedron to render high-frequency details.
\vspace{-0.175em}

\begin{figure}[t]
\centering
\begin{overpic}[width=\linewidth]{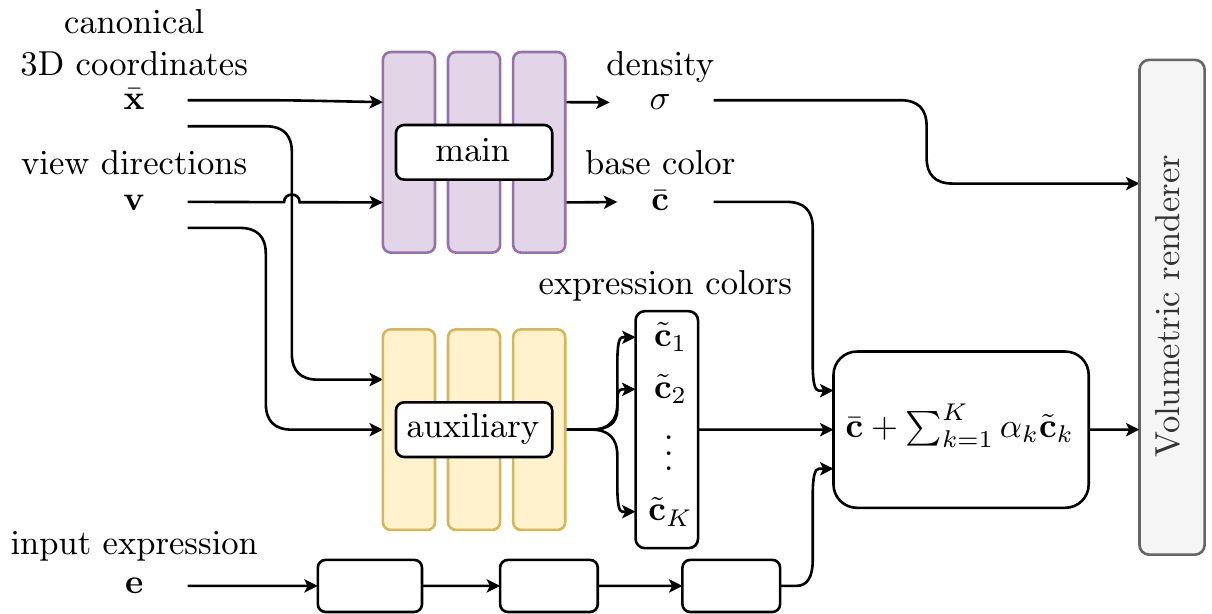}
    \put (26.9, 1.4) {%
        \scalebox{0.65}{Eq.~\ref{eq:local-descriptor}}%
    }
    \put (41.8, 1.4) {%
        \scalebox{0.65}{Eq.~\ref{eq:local-similarity}}%
    }
    \put (56.7, 1.4) {%
        \scalebox{0.65}{Eq.~\ref{eq:blendfield}}%
    }
\end{overpic}

\caption{
\textbf{\methodname{} } -- 
We implement 
{our approach as a volumetric model},
 where the \textit{appearance} (i.e. radiance) is the sum of the main appearance corrected by blending a small set of $\nExpr$ expression-specific appearances. These appearances are learnt from extreme expressions, and then blended at test-time according to blend weights computed as a function of the input expression~$\expression$.
}
\label{fig:pipeline}
\vspace{-1.25em}
\end{figure}

\section{Method}
\label{ref:method}

We introduce a volumetric model that can be driven by input expressions and visualize it in in~\cref{fig:pipeline}. We start this section by explaining our model and how we train and drive it with novel expressions utilizing parametric face models (\cref{sec:model}).
We then discuss how to compute measures of volume expansion and compression in the tetrahedra to combine volumetric models of different expressions (\cref{sec:similarity}) and how we remove artifacts in out-of-distribution settings (\cref{sec:smoothness}).
We conclude this section with implementation details (\cref{sec:implementation}).

\subsection{Our model}
\label{sec:model}
Given a neutral expression $\template\expression$, and a collection of posed images $\{\pixelcolor_c\}$ of this expression from multiple views, \VolTeMorph employs a map~$\map$ to fetch the density and radiance\footnote{We omit view-dependent effects to simplify notation but include them in our implementation.} for a new expression $\expression$ from the \textit{canonical} frame defined by expression $\template\expression$:
\begin{align}
\radiance(\pos; \expression) &= \template\radiance(\bar\pos), \quad \bar\pos = \map(\pos; \expression \rightarrow \template\expression)
\\
\density(\pos; \expression) &= \template\density(\bar\pos), \quad \bar\pos = \map(\pos; \expression \rightarrow \template\expression)
\label{eq:densityVoltemorph}
\\
\loss_\text{rgb} &= 
\expect_{\pixelcolor \sim \{\pixelcolor_c\}} \:
\expect_{\ray \sim \pixelcolor} \:
\loss_\text{rgb}^\ray
\\
\loss_\text{rgb}^\ray &= \| \pixelcolor(\ray; \expression) - \pixelcolor(\ray) \|_2^2,
\end{align}
where $\pixelcolor(\ray; \expression)$ is a pixel color produced by our model conditioned on the input expression $\expression$, $\pixelcolor(\ray)$ is the ground-truth pixel color, and the mapping~$\map$ is computed from smooth deformations of a tetrahedral mesh to render unseen expressions~$\expression$. 
We use expression vectors $\expression$ from parametric face models, such as FLAME~\cite{li2017flame,fakeItTillYouMakeIt}.
However, as neither density nor radiance change with $\expression$, changes in appearance are limited to the low-frequency deformations that $\map$ can express.
For example, this model cannot capture high-frequency dynamic features like expression wrinkles.
We overcome this limitation by conditioning radiance on expression. For this purpose, we assume radiance to be the sum of a template radiance (\ie rest pose appearance of a subject) and $\nExpr$ residual radiances (\ie details belonging to corresponding facial expressions):
\begin{equation}
\outputcolor(\pos; \expression) = \bar\radiance(\pos) + \sum_{k=\iExpr}^\nExpr
\blendingweight_\iExpr(\pos; \expression) 
\cdot
\aux\radiance_\iExpr(\pos), 
\label{eq:multiheadradiance}
\end{equation}
We call our model \textit{blend fields}, as it resembles the way in which blending is employed in 3D morphable models~\cite{blanz1999morphable} or in wrinkle maps~\cite{oat2007animated}.
Note that we assume that pose-dependent geometry can be effectively modeled as 
{a convex combination of colors $[\aux\radiance(\pos)]_{\iExpr=1}^{\nExpr}$}, since we employ the same density fields as in~\eq{densityVoltemorph}. %
In what follows, for convenience, we denote the vector field of blending coefficients as $\blendfield(\pos) {=} [\blendingweight_\iExpr(\pos)]_{\iExpr=1}^{\nExpr}$.

\renewcommand{\imagewidth}{0.25\linewidth}
\renewcommand{\smallimagewidth}{0.12\linewidth}
\newcommand{\offset}{0.8pt}
\begin{figure}[t]
    \centering
        \begin{subfigure}[b]{\linewidth}
            \centering
            \resizebox{\linewidth}{!}{
                \begin{tikzpicture}[
                    >=stealth',
                    overlay/.style={
                      anchor=north west, 
                      draw=black,
                      rectangle, 
                      line width=\offset,
                      outer sep=0,
                      inner sep=0,
                    },
                    every node/.style={inner sep=0,outer sep=0}
                ]
                    \matrix[matrix of nodes, column sep=-1pt, row sep=0, ampersand replacement=\&] (datarep) {
                        \includegraphics[width=\imagewidth,trim={1.2cm 0 0 0},clip]{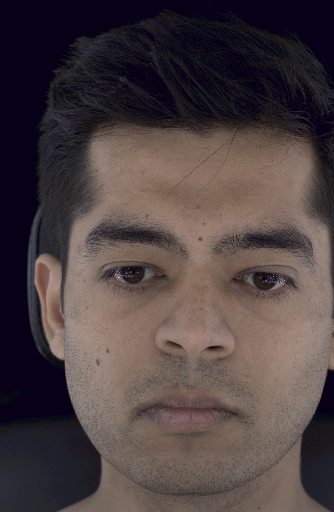} \&
                        \includegraphics[width=\imagewidth,trim={1.2cm 0 0 0},clip]{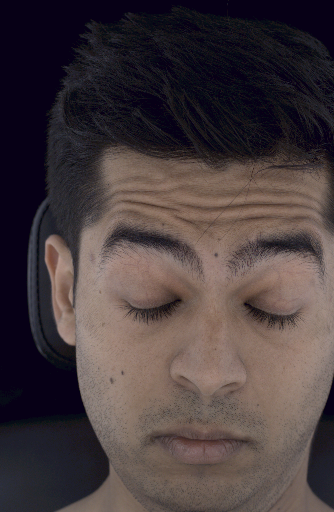} \&
                        \includegraphics[width=\imagewidth,trim={1.2cm 0 0 0},clip]{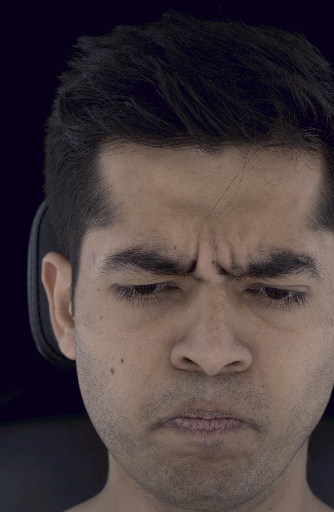} \&
                        \includegraphics[width=\imagewidth,trim={1.2cm 0 0 0},clip]{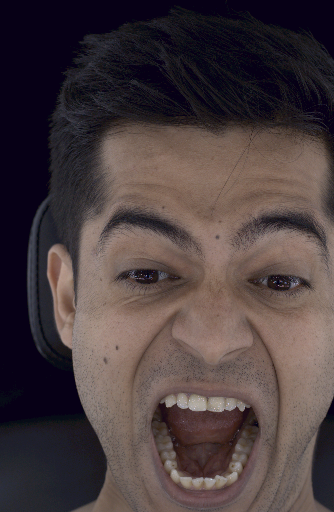} \\ %
                    };
                    \node[below right=\offset and \offset of datarep-1-1.north west, overlay]{
                        \includegraphics[width=\smallimagewidth]{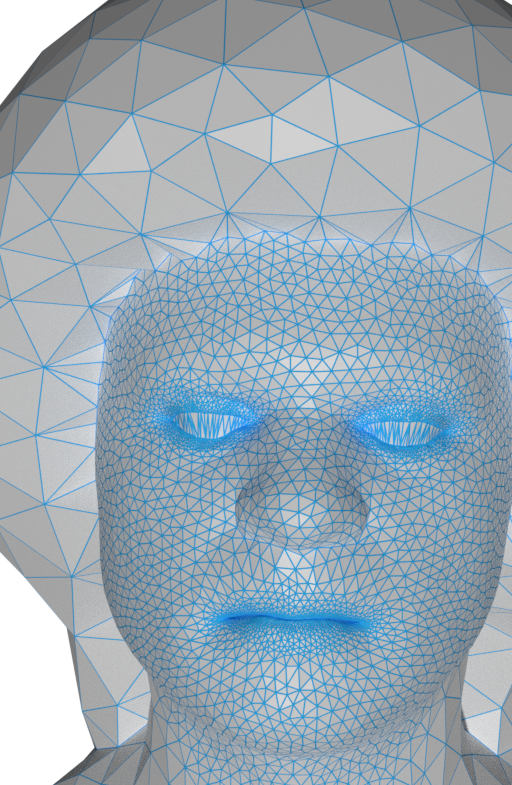}
                    };
                    \node[below right=\offset and \offset of datarep-1-2.north west, overlay]{
                        \includegraphics[width=\smallimagewidth]{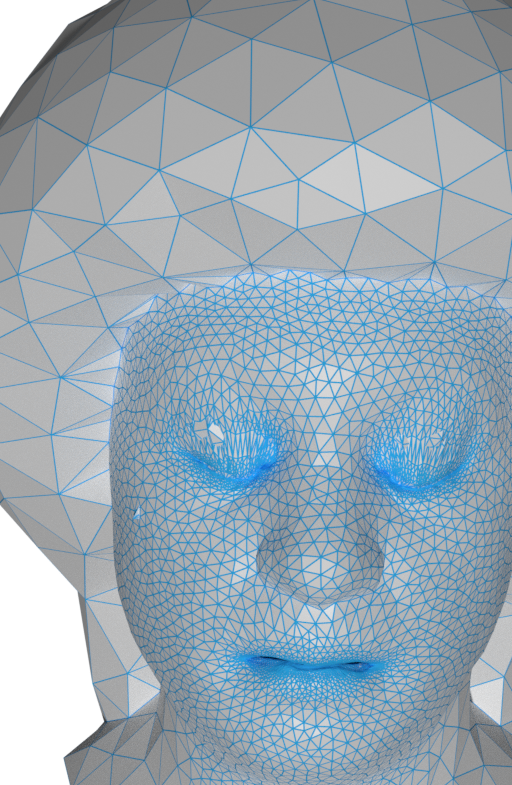}
                    };
                    \node[below right=\offset and \offset of datarep-1-3.north west, overlay]{
                        \includegraphics[width=\smallimagewidth]{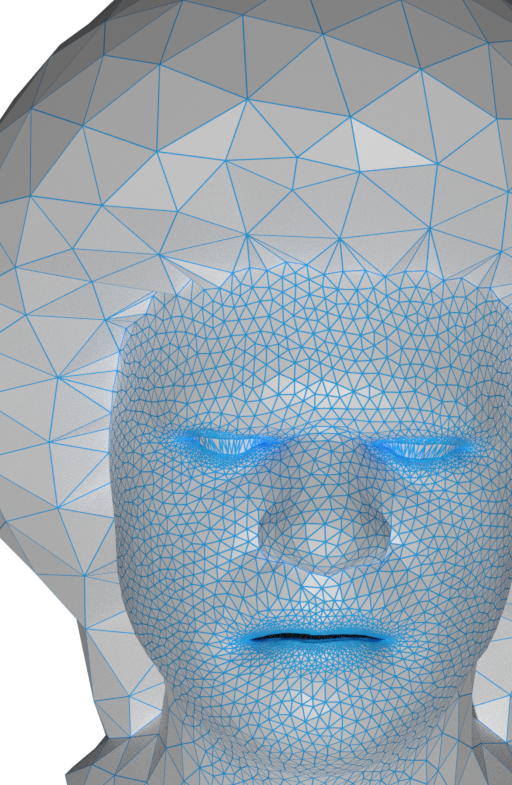}
                    };
                    \node[below right=\offset and \offset of datarep-1-4.north west, overlay]{
                        \includegraphics[width=\smallimagewidth]{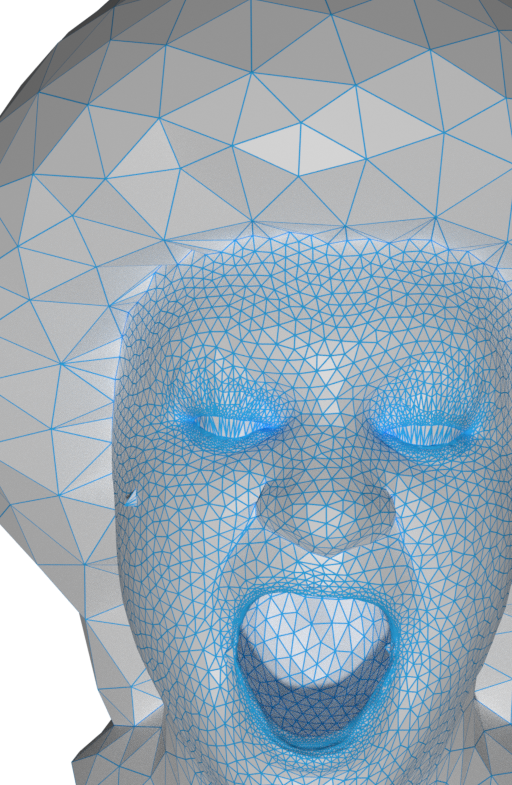}
                    };
                \end{tikzpicture}
            }
        \end{subfigure}
    \caption{\textbf{Data} -- We represent the data as a multi-view, multi-expression images. For each of these images, we obtain parameters of a parametric model, such as FLAME \cite{li2017flame} to get: an expression vector $\expression$ and a tetrahedral mesh described by vertices $\meshVertices(\expression)$. We highlight that our approach works for any object if a rough mesh and its descriptor are already provided.
    }
    \label{fig:data-representation}
\vspace{-1.25em}
\end{figure}
\noindent\textbf{Training the model}
We train our model by assuming that we have access to a small set of $\nExpr$ images~$\{\image_\iExpr\}$ (example in \cref{fig:data-representation}), each corresponding to an ``extreme'' expression $\{\expression_\iExpr\}$, and minimize the loss:
\begin{align}
\label{eq:loss}
\loss_\text{rgb} &= 
\expect_{\iExpr} \:
\expect_{\ray} \:
\| \pixelcolor_\nExpr(\ray; \expression_\iExpr) - \pixelcolor_\iExpr(\ray) \|_2^2 
\\
&\text{where} \quad \forall \pos,\:\: \blendfield(\pos) = \indicator_\iExpr,
\end{align}
where $\indicator_\iExpr$ is the indicator vector, which has value one at the $k$-th position and zeroes elsewhere, and $\pixelcolor_\nExpr$ represents the output of integrating the radiances in~\eq{multiheadradiance} along a ray.

\noindent\textbf{Driving the model}
To control our model given a novel expression $\expression$, we need to map the input expression code to the corresponding blendfield~$\blendfield(\pos)$.
We parameterize the blend field as a vector field discretized on the vertices $\meshVertices(\expression)$ of our tetrahedral mesh, where the vertices deform according to the given expression. The field is discretized on vertices, but it can be queried within tetrahedra using linear FEM bases~\cite{monk2003finite}. 
Our core intuition is that when the~(local) geometry of the mesh matches the local geometry in one of the input expressions, the corresponding expression blend weight should be locally activated. More formally, let $\vertex{\in} \meshVertices$ be a vertex in the tetrahedra and $\geometry(\vertex)$ a local measure of volume on the vertex described in~\cref{sec:similarity}, then
\begin{align}
\geometry(\vertex(\expression)) {\approx} \geometry(\vertex(\expression_\iExpr))
\Longrightarrow
\blendfield(\vertex(\expression)) \approx \indicator_\iExpr.
\label{eq:coreidea}
\end{align}
To achieve this we first define a \textit{local} similarity measure:
\begin{equation}
\label{eq:local-similarity}
[ \Delta\geometry_k(\vertex(\expression))] = [ \| \geometry(\vertex(\expression)) {-} \geometry(\vertex(\expression_\iExpr))\|_2^2 ] \in \real^{\nExpr}
\end{equation}
and then gate it with softmax (with temperature $\temp{=}10^6$) to obtain vertex blend weights:
\begin{align}
\label{eq:blendfield}
\blendfield(\vertex(\expression)) = \text{softmax}_\temp \{ \Delta\geometry_k(\vertex(\expression)) \} \in [0,1]^{\nExpr}
\end{align}
which realizes~\eq{coreidea}, as well as preserves the typically desirable characteristics of blend weights:
\begin{itemize}
\item \textit{partition of unity}: $\forall\pos \:\: \blendfield(\pos) \in [0,1]^\nExpr$ and $\| \blendfield(\pos) \|_1{=}1$ 
\item \textit{activations sparsity}: minimizers of $\|\blendfield(\pos)\|_0$
\end{itemize}
where the former ensures any reconstructed result is a~\textit{convex} combination of input data, and the latter prevents destructive interference~\cite{ichim2015dynamic}.

\subsection{Local geometry descriptor}
\label{sec:similarity}
Let us consider a tetrahedron as the matrix formed by its vertices~$\tet {=} \{\vertex_i\} \in \real^{3 \times 4}$, and its edge matrix as $\edgematrix = [\vertex_3-\vertex_0, \vertex_2-\vertex_0, \vertex_1-\vertex_0]$.
Let us denote $\template\edgematrix$ as the edge matrix in rest pose and $\edgematrix$ as one of the deformed tetrahedra (\ie, due to expression). 
From classical FEM literature, we can then compute the change in volume of the tetrahedra from the determinant of its deformation gradient~\cite{irving2004invertible}:
\begin{equation}
\label{eq:volume}
\Delta \volume(\tet) = \text{det}(\edgematrix \cdot \template\edgematrix^{-1})
\end{equation}
We then build a local volumetric descriptor for a specific (deformed) vertex $\vertex(\expression)$ by concatenating the changes in volumes of neighboring (deformed) tetrahedra:
\begin{align}
\label{eq:local-descriptor}
\geometry(\vertex(\expression)) =
\bigoplus_{\tet \in \neighbourhood(\vertex)}
\Delta \volume(\tet(\expression)),
\end{align}
where $\bigoplus$ denotes concatenation and $\neighbourhood(\vertex)$ topological neighborhood of a vertex $\vertex$.  

\begin{figure}[t]
    \centering
    \begin{overpic}[width=\linewidth, trim=0 0 -1.3em -1em, clip]{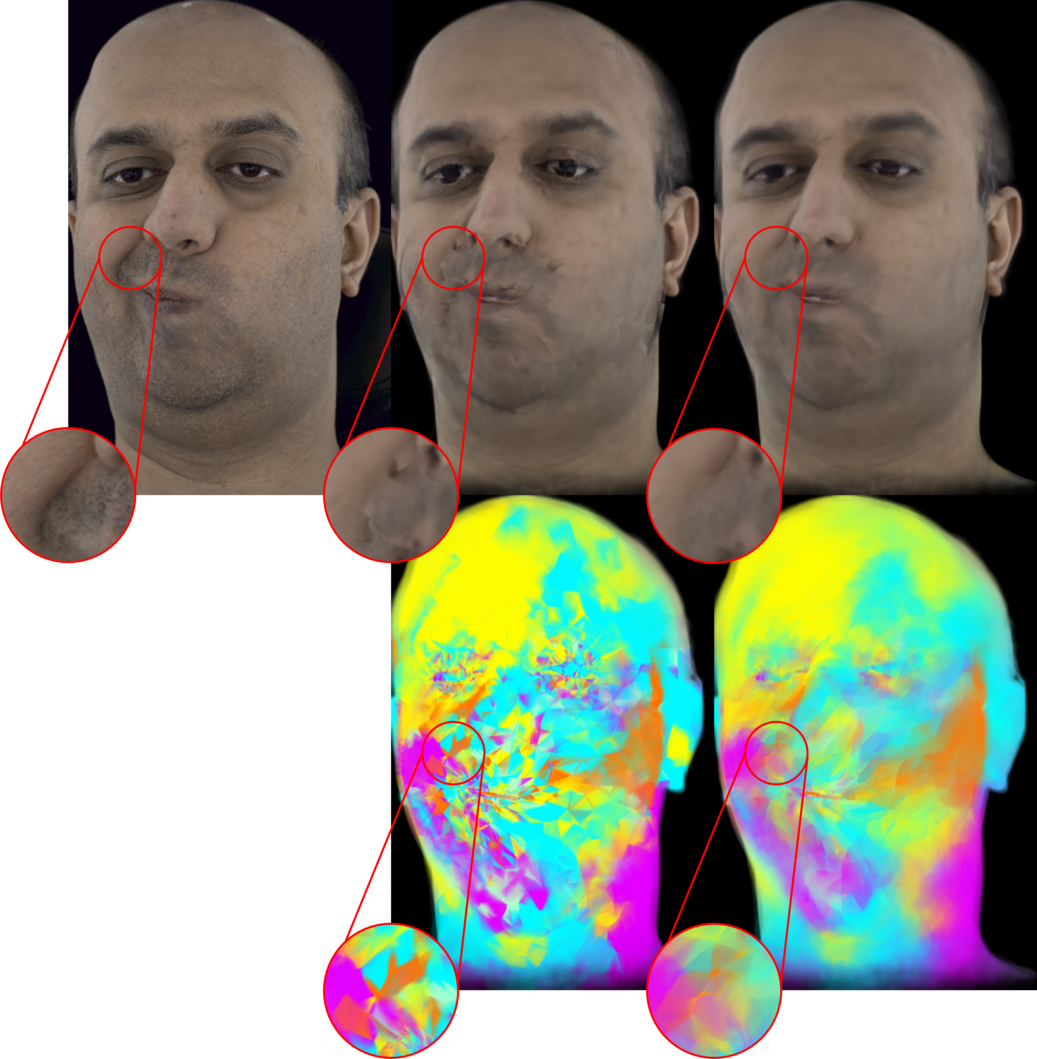}
        \put(11,98) {GT image}
        \put(37,98) {No smoothing}
        \put(65,98) {With smoothing}
        \put(95,79) {\rotatebox{-90}{RGB}}
        \put(95,46) {\rotatebox{-90}{$\blendfield(\pos)$ space (\cref{eq:blendfield})}}
    \end{overpic}
    \vspace{-2em}
    \caption{\textbf{Laplacian smoothing} -- To combat artifacts stemming from calculating weights $\blendingweight$ across multiple expressions, which may assign different expressions to neighboring tetrahedra, we apply Laplacian smoothing~\cite{desbrun1999implicit}. As seen in the bottom row, smoothing gives a more consistent expression assignment.}
    \label{fig:laplacian-smoothing}
\vspace{-1.25em}
\end{figure}

\renewcommand{\imagewidth}{2.3cm}
\renewcommand{\imageheight}{4.0263671874999996cm}
\renewcommand{\smallimagewidth}{0.8cm}
\newcommand{\expressionsep}{0em}

\newcommand{\imageoneindex}[2]{%
    \includegraphics[height=\imageheight, trim={0 0 0 0}, clip]{assets/qualitative/main/#1_6795937_extrapolation_#2.png}%
}
\newcommand{\imagetwoindex}[2]{%
    \includegraphics[height=\imageheight, trim={0 0 0 0}, clip]{assets/qualitative/main/#1_7889059_extrapolation_#2.png}%
}
\renewcommand{\versionone}{
    \begin{tikzpicture}[
        >=stealth',
        overlay/.style={
          anchor=south west, 
          draw=black,
          rectangle, 
          line width=0.8pt,
          outer sep=0,
          inner sep=0,
        },
    ]
        \matrix[
        matrix of nodes, 
        column sep=0pt, 
        row sep=0pt, 
        ampersand replacement=\&, 
        inner sep=0, 
        outer sep=0
        ] (gt) {
            \imageoneindex{gt}{0505} \&
            \imageoneindex{gt}{0000} \&
            \imagetwoindex{gt}{0000} \&
            \imagetwoindex{gt}{0852} \\
        };
        \matrix[
        matrix of nodes, 
        column sep=0pt, 
        row sep=0pt, 
        ampersand replacement=\&, 
        inner sep=0, 
        outer sep=0,
        below=0pt of gt,
        ] (voltemorphstatic) {
            \imageoneindex{voltemorph_static}{0505} \&
            \imageoneindex{voltemorph_static}{0000} \&
            \imagetwoindex{voltemorph_static}{0000} \&
            \imagetwoindex{voltemorph_static}{0852} \\
        };
        
        \matrix[
        matrix of nodes, 
        column sep=0pt, 
        row sep=0pt, 
        ampersand replacement=\&, 
        inner sep=0, 
        outer sep=0,
        below=0pt of voltemorphstatic
        ] (voltemorph) {
            \imageoneindex{voltemorph}{0505} \&
            \imageoneindex{voltemorph}{0000} \&
            \imagetwoindex{voltemorph}{0000} \&
            \imagetwoindex{voltemorph}{0852} \\
        };
        
        \matrix[
        matrix of nodes, 
        column sep=0pt, 
        row sep=0pt, 
        ampersand replacement=\&, 
        inner sep=0, 
        outer sep=0,
        below=0pt of voltemorph
        ] (ours) {
            \imageoneindex{bv_aux}{0505} \&
            \imageoneindex{bv_aux}{0000} \&
            \imagetwoindex{bv_aux}{0000} \&
            \imagetwoindex{bv_aux}{0852} \\
        };

        \node[left=0em of gt-1-1.west, align=center, anchor=east]{\rotatebox{90}{Ground Truth}};
        \node[left=0em of voltemorphstatic-1-1.west, align=center, anchor=east]{\rotatebox{90}{VolTeMorph$_1$}};
        \node[left=0em of voltemorph-1-1.west, align=center, anchor=east]{\rotatebox{90}{VolTeMorph$_\text{avg}$}};
        \node[left=0em of ours-1-1.west, align=center, anchor=east]{\rotatebox{90}{\textbf{Ours}}};
        
        \node[above=0em of gt-1-1.north, align=center, anchor=south]{Neutral};
        \node[above=0em of gt-1-2.north, align=center, anchor=south]{Posed};
        \node[above=0em of gt-1-3.north, align=center, anchor=south]{Neutral};
        \node[above=0em of gt-1-4.north, align=center, anchor=south]{Posed};
    \end{tikzpicture}
}

\begin{figure*}[htb]
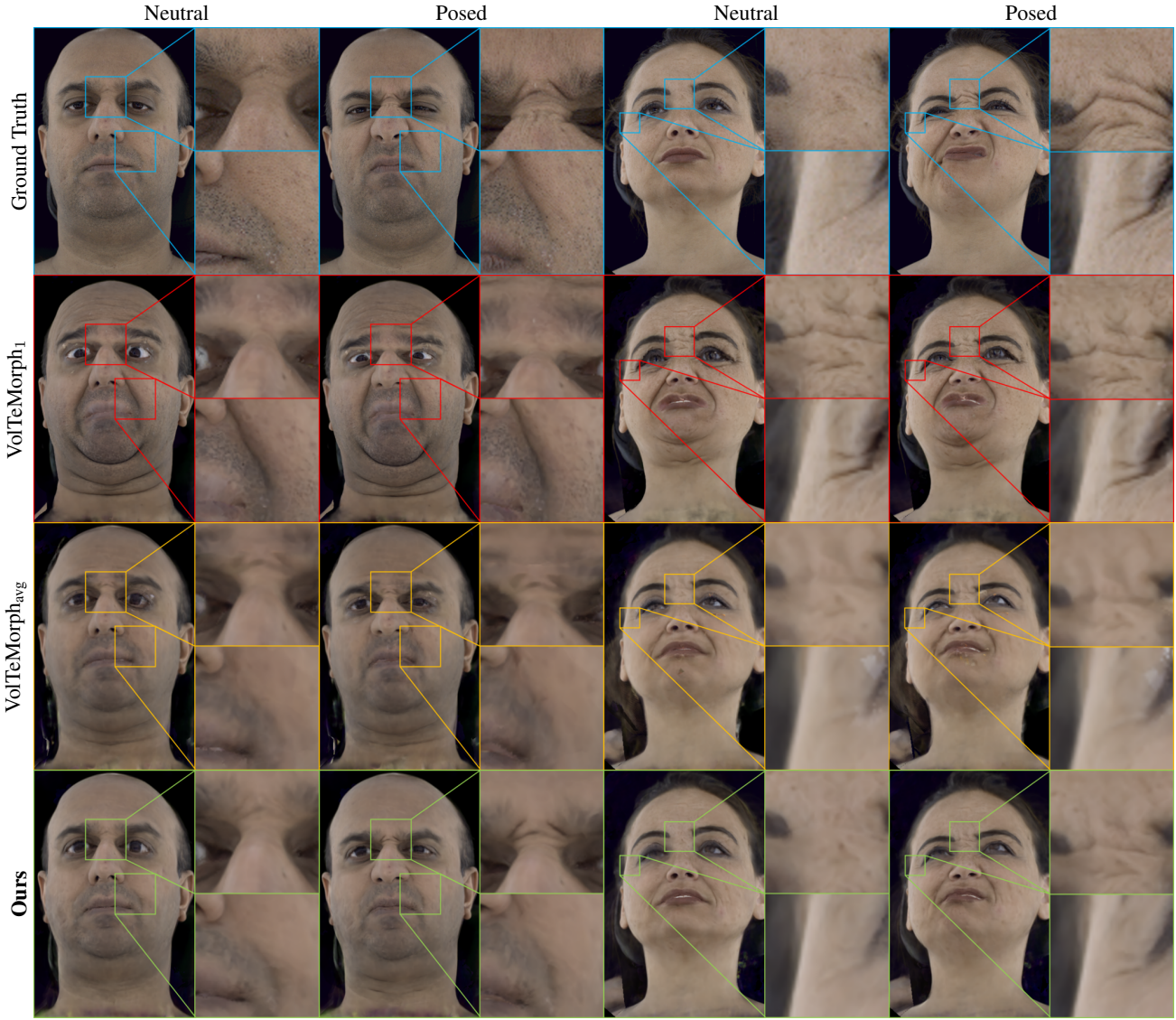

    \centering
    \resizebox{\linewidth}{!}{\versionone}
    \caption{\textbf{Novel expression synthesis} -- 
    We compare qualitatively \methodname{} with selected baselines (vertical) across two selected subjects (horizontal). Firstly, we show a neutral pose of the subject and then any of the available expressions. To our surprise, VolTeMorph$_\text{avg}$ trained on multiple frames renders some details but with much lower fidelity. We argue that VolTeMorph$_\text{arg}$ considers rendering wrinkles as artifacts that depend on the view direction (see~\Cref{eq:rendering-equation}). VolTeMorph$_1$ is limited to producing the wrinkles it was trained for. In contrast to those baselines, \textbf{\methodname{}} captures the details and generalizes outside of the distribution. Please refer to the \supplementary{} for animated sequences and results for other methods. %
    }
    \vspace{-1.25em}
    \label{fig:qualitative-comparison}
\end{figure*}
\subsection{Blend-field smoothness}
\label{sec:smoothness}
\newcommand{\blendfieldVertices}{\mathbf{A}}
High-frequency spatial changes in blendfields can cause visual artifacts, see~\cref{fig:laplacian-smoothing}. We overcome this issue by applying a small amount of smoothing to the blendfield.
Let us denote with $\blendfieldVertices{=}\{\blendfield(\vertex_v)\}$ the matrix of blend fields defined on all mesh vertices, and with $\mathbf{L}$ the Laplace-Beltrami operator for the tetrahedral mesh induced by linear bases~\cite{irving2004invertible}.
We exploit the fact that at test-time, the field is discretized on the mesh vertices, execute a diffusion process on the tetrahedral manifold, and, to avoid instability problems, implement it via backward Euler~\cite{desbrun1999implicit}:
\begin{equation}
\blendfieldVertices^\text{diff} = (\mathbf{I} - \hyperparam_\text{diff} \laplacian)^{-1} \blendfieldVertices^n.
\end{equation}

\subsection{Implementation details}
\label{sec:implementation}
We build on VolTeMorph~\cite{garbin2022voltemorph} and use its volumetric 3DMM face model. However, the same methodology can be used with other tetrahedral cages built on top of 3DMM face models. The face model is created by extending the blendshapes of the parametric 3DMM face model~\cite{fakeItTillYouMakeIt} to a tetrahedral cage that defines the support in the neural radiance field. It has four bones controlling global rotation, the neck and the eyes with linear blend skinning, 224 expression blendshapes, and 256 identity blendshapes. Our face radiance fields are thus controlled and posed with the identity, expression, and pose parameters of the 3DMM face model~\cite{fakeItTillYouMakeIt}, can be estimated by a real-time face tracking system like~\cite{wft}, and generalize convincingly to expressions representable by the face model.

\noindent\textbf{Training.} During training, we sample rays from a single frame to avoid out-of-memory issues when evaluating the tetrahedral mesh for multiple frames. Each batch contains~1024 rays. We sample $\coarsesamples{=}128$ points along a single ray during the coarse sampling and $\importancesamples{=}64$ for the importance sampling. We train the network to minimize the loss in~\cref{eq:loss} and sparsity losses with standard weights used in VolTeMorph~\cite{garbin2022voltemorph,hedman2021baking}. We train the methods for~$5{\times}10^5$ steps using Adam~\cite{kingma2014adam} optimizer with learning rate~$5{\times}10^{-4}$ decaying exponentially by factor of~$0.1$ every~$5{\times}10^5$ steps.

\noindent\textbf{Inference.} During inference, we leverage the underlying mesh to sample points around tetrahedra hit by a single ray. Therefore, we perform a single-stage sampling with $\numberofsamples{=}\coarsesamples{+}\importancesamples$ samples along the ray. When extracting the features~(\cref{eq:local-descriptor}), we consider $|\neighbourhood(\vertex)|{=}20$ neighbors. 
For the Laplacian smoothing, we set $\weightdiffusion{=}0.1$ and perform a single iteration step.
Geometric-related operations impose negligible computational overhead.

\begin{table*}[!t]
    \centering
    \resizebox{\linewidth}{!}{
        \begin{tabular}{lccccccccc}
        \toprule
        \multirow{3}[3]{*}{Method} & \multicolumn{6}{c}{Real Data} &\multicolumn{3}{c}{Synthetic Data}\\
        \cmidrule(lr){2-7}\cmidrule(lr){8-10}
         & \multicolumn{3}{c}{Casual Expressions} & \multicolumn{3}{c}{Novel Pose Synthesis}& \multicolumn{3}{c}{Novel Pose Synthesis} \\
        \cmidrule(lr){2-4}\cmidrule(lr){5-7}\cmidrule(lr){8-10}
        & PSNR $\uparrow$ & SSIM $\uparrow$ & LPIPS $\downarrow$ & PSNR $\uparrow$ & SSIM $\uparrow$ & LPIPS $\downarrow$& PSNR $\uparrow$ & SSIM $\uparrow$ & LPIPS $\downarrow$ \\
        \midrule
NeRF \cite{mildenhall2020nerf}                     &                             23.6465 &                             0.7384 &                             0.2209  &                             25.6696 &                             0.8127 &                             0.1861 &                             13.7210 &                             0.6868 &                             0.3113 \\
Conditioned NeRF \cite{mildenhall2020nerf}         &                             22.9106 &                             0.7162 &                             0.2029  &                             24.7283 &                             0.7927 &                             0.1682 &                             19.5971 &                             0.8138 &                             0.1545 \\
NeRFies \cite{park2021nerfies}                     &                             22.6571 &                             0.7105 &                             0.2271  &                             24.8376 &                             0.7990 &                             0.1884 &                             19.3042 &                             0.8081 &                             0.1591 \\
HyperNeRF-AP \cite{park2021hypernerf}              &                             22.6219 &                             0.7087 &                             0.2236  &                             24.7119 &                             0.7931 &                             0.1848 &                             19.3557 &                             0.8132 &                             0.1563 \\
HyperNeRF-DS \cite{park2021hypernerf}              &                             22.9299 &                             0.7182 &                             0.2241  &                             24.9909 &                             0.8007 &                             0.1860 &                             19.4637 &                             0.8159 &                             0.1526 \\
\midrule
VolTeMorph$_1$ \cite{garbin2022voltemorph}         &                             24.9939 &                             0.8358 &                             0.1164  &                             26.7526 &                             0.8749 &  \cellcolor{secondbestcolor}0.0954 &                             26.7033 &                             0.9500 &                             0.0394 \\
VolTeMorph$_\text{avg}$\cite{garbin2022voltemorph} &  \cellcolor{secondbestcolor}26.9209 &  \cellcolor{secondbestcolor}0.8912 &  \cellcolor{secondbestcolor}0.1105  &  \cellcolor{secondbestcolor}28.6866 &  \cellcolor{secondbestcolor}0.9176 &                             0.0982 &  \cellcolor{secondbestcolor}30.2107 &  \cellcolor{secondbestcolor}0.9815 &  \cellcolor{secondbestcolor}0.0387 \\
\midrule
\textbf{\methodname{}}                             &   \cellcolor{firstbestcolor}27.5977 &   \cellcolor{firstbestcolor}0.9056 &   \cellcolor{firstbestcolor}0.0854  &   \cellcolor{firstbestcolor}29.7372 &   \cellcolor{firstbestcolor}0.9311 &   \cellcolor{firstbestcolor}0.0782 &   \cellcolor{firstbestcolor}32.7949 &   \cellcolor{firstbestcolor}0.9882 &   \cellcolor{firstbestcolor}0.0221 \\
\bottomrule  
        \end{tabular}
    }
    \caption{\textbf{Quantitative results} -- 
    We compare \methodname{} to other related approaches. We split the real data into two settings: one with casual expressions of subjects and the other with novel, static expressions.
    For the real data, we only compute metrics on the face region, which we separate using an off-the-shelf face segmentation network~\cite{wood2021fake}.
    Please refer to the \supplementary{} for the results that include the background in the metrics as well.
    We average results across frames and subjects. 
    VolTeMorph$_{\text{avg}}$~\cite{garbin2022voltemorph} is trained on all frames, while VolTeMorph$_1$ is trained on a single frame. HyperNeRF-AP/-DS follows the design principles from Park~\etal~\cite{park2021hypernerf}. The best results are colored in \mycoloredbox{firstbestcolor} and second best results in \mycoloredbox{secondbestcolor}. \methodname{} performs best in most of the datasets and metrics. Please note that HyperNeRF-AP/DS and NeRFies predict a dense deformation field designed for dense data. However, our input data consists of a few static frames only where the deformation field leads to severe overfitting.}
    \label{tab:quantitative-results}
\end{table*}

\begin{table*}[!t]
    \centering
        \begin{tabular}{lccccccccc}
        \toprule
        \multirow{3}[3]{*}{Parameter} & \multicolumn{6}{c}{Real Data} &\multicolumn{3}{c}{Synthetic Data}\\
        \cmidrule(lr){2-7}\cmidrule(lr){8-10}
         & \multicolumn{3}{c}{Casual Expressions} & \multicolumn{3}{c}{Novel Pose Synthesis}& \multicolumn{3}{c}{Novel Pose Synthesis} \\
        \cmidrule(lr){2-4}\cmidrule(lr){5-7}\cmidrule(lr){8-10}
        & PSNR $\uparrow$ & SSIM $\uparrow$ & LPIPS $\downarrow$ & PSNR $\uparrow$ & SSIM $\uparrow$ & LPIPS $\downarrow$& PSNR $\uparrow$ & SSIM $\uparrow$ & LPIPS $\downarrow$ \\
        \midrule
        $|\neighbourhood(\vertex)| = 1$  &                            27.5620 &                             0.9043 &                             0.0893 &                             29.7269 &                             0.9306 &                             0.0815 &                             32.2371 &  \cellcolor{secondbestcolor}0.9882 &                             0.0234   \\
        $|\neighbourhood(\vertex)| = 5$  &                            27.5880 &  \cellcolor{secondbestcolor}0.9054 &                             0.0864 &   \cellcolor{firstbestcolor}29.7548 &  \cellcolor{secondbestcolor}0.9312 &                             0.0789 &                             32.2900 &  \cellcolor{secondbestcolor}0.9882 &                             0.0231   \\
        $|\neighbourhood(\vertex)| = 10$ & \cellcolor{secondbestcolor}27.5933 &  \cellcolor{secondbestcolor}0.9054 &  \cellcolor{secondbestcolor}0.0859 &  \cellcolor{secondbestcolor}29.7456 &  \cellcolor{secondbestcolor}0.9312 &  \cellcolor{secondbestcolor}0.0785 &  \cellcolor{secondbestcolor}32.3324 &  \cellcolor{secondbestcolor}0.9882 &  \cellcolor{secondbestcolor}0.0230   \\
        $|\neighbourhood(\vertex)| = 20$ &  \cellcolor{firstbestcolor}27.5977 &   \cellcolor{firstbestcolor}0.9056 &   \cellcolor{firstbestcolor}0.0854 &                             29.7372 &                             0.9311 &   \cellcolor{firstbestcolor}0.0782 &   \cellcolor{firstbestcolor}32.7949 &   \cellcolor{firstbestcolor}0.9887 &   \cellcolor{firstbestcolor}0.0221   \\
        \midrule                         
        Without smoothing    & \cellcolor{secondbestcolor}27.2535 &  \cellcolor{secondbestcolor}0.8959 &  \cellcolor{secondbestcolor}0.0939  &  \cellcolor{secondbestcolor}29.3726 &  \cellcolor{secondbestcolor}0.9233 &  \cellcolor{secondbestcolor}0.0846&  \cellcolor{secondbestcolor}32.2452 &  \cellcolor{secondbestcolor}0.9876 &  \cellcolor{secondbestcolor}0.0238 \\
        With smoothing       &  \cellcolor{firstbestcolor}27.5977 &   \cellcolor{firstbestcolor}0.9056 &   \cellcolor{firstbestcolor}0.0854  &   \cellcolor{firstbestcolor}29.7372 &   \cellcolor{firstbestcolor}0.9311 &   \cellcolor{firstbestcolor}0.0782&   \cellcolor{firstbestcolor}32.7949 &   \cellcolor{firstbestcolor}0.9887 &   \cellcolor{firstbestcolor}0.0221 \\
        \bottomrule
        \end{tabular}
    \caption{\textbf{Ablation study} -- {
            First, we check the effect of the neighborhood size $|\neighbourhood(\vertex)|$ on the results. Below that, we compare the effect of smoothing. 
            The best results are colored in \mycoloredbox{firstbestcolor} and the second best in \mycoloredbox{secondbestcolor}. For the real dataset, changing the neighborhood size gives inconsistent results, while smoothing improves the rendering quality. In the synthetic scenario, setting $|\neighbourhood(\vertex)|{=}20$ and the Laplacian smoothing consistently gives the best results. The discrepancy between real and synthetic datasets is caused by inaccurate face tracking for the former. We describe this issue in detail in~\Cref{subsec:failures}.
        }
    }
    \label{tab:ablation-study}
    \vspace{-1.2em}
\end{table*}

\renewcommand{\imagewidth}{2.3cm}
\renewcommand{\imageheight}{4.0cm}
\renewcommand{\smallimagewidth}{0.8cm}
\newcommand{\borderwidth}{2.0pt}
\newcommand{\spacingbetweencols}{2.0pt}

\newcommand{\imageindex}[2]{%
    \includegraphics[height=\imageheight, trim={4.6cm 1.5cm 4.6cm 1.5cm}, clip]{assets/synthetics/#1_synthetic_#2}%
}
\renewcommand{\versionone}{
    \begin{tikzpicture}[
        >=stealth',
        overlay/.style={
          anchor=south west, 
          draw=black,
          rectangle, 
          line width=0.8pt,
          outer sep=0,
          inner sep=0,
        },
    ]
        \matrix[
        matrix of nodes, 
        column sep=0pt, 
        row sep=0pt, 
        ampersand replacement=\&, 
        inner sep=0, 
        outer sep=0,
        ] (main) {
            \imageindex{gt}{0000} \\
            \imageindex{gt}{0002} \\
        };
        \matrix[
        matrix of nodes, 
        column sep=0pt, 
        row sep=0pt, 
        ampersand replacement=\&, 
        inner sep=0, 
        outer sep=0,
        right=\spacingbetweencols of main
        ] (voltemorph) {
            \imageindex{voltemorph_static}{0000} \&
            \imageindex{voltemorph}{0000} \\
            \imageindex{voltemorph_static}{0002} \&
            \imageindex{voltemorph}{0002} \\
        };
        \matrix[
        matrix of nodes, 
        column sep=0pt, 
        row sep=0pt, 
        ampersand replacement=\&, 
        inner sep=0, 
        outer sep=0,
        right=\spacingbetweencols of voltemorph
        ] (blendvolumes) {
            \imageindex{blendvolumes_aux}{0000} \\
            \imageindex{blendvolumes_aux}{0002} \\
        };
        
        \node[above=0.2em of main-1-1.north, align=center, anchor=south]{Ground Truth};
        \node[above=0.0em of voltemorph-1-1.north, align=center, anchor=south]{VolTeMorph$_1$};
        \node[above=0.0em of voltemorph-1-2.north, align=center, anchor=south]{VolTeMorph$_\text{avg}$};
        \node[above=0.1em of blendvolumes-1-1.north, align=center, anchor=south]{\textbf{Ours}};
        
        \node[left=0.1em of main-1-1.west, align=center, anchor=east]{\rotatebox{90}{Canonical}};
        \node[left=0.1em of main-2-1.west, align=center, anchor=east]{\rotatebox{90}{Twisted}};
    \end{tikzpicture}
}
\begin{figure}[t]
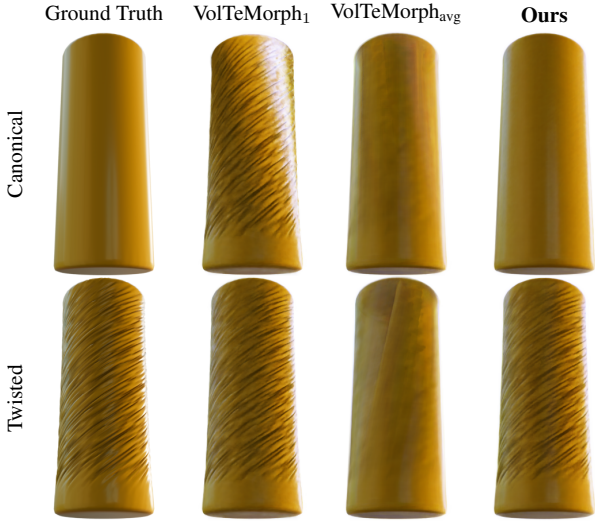

    \centering
    \resizebox{\linewidth}{!}{\versionone}
    \caption{\textbf{Qualitative results on synthetic dataset} --  For a simple dataset, baselines cannot model high-frequency, pose-dependent details. VolTeMorph$_1$ renders wrinkles for the straight pose as well, as it is trained for the twisted cylinder only, while VolTeMorph$_\text{avg}$ averages out the texture.}
    \label{fig:synthetic-qualitative}
    \vspace{-1.25em}
\end{figure}

\section{Experiments}
\label{ref:experiments}

We evaluate all methods on data of four subjects from the publicly available Multiface dataset~\cite{wuu2022multiface}.
We track the face 
for eight manually-selected ''extreme'' expressions.
We then select $\nExpr{=}5$ expressions 
the combinations of which show
 as many wrinkles as possible.
Each subject
was captured with $\approx\!\!38$ cameras which gives~$\approx\!\!190$ training images per subject%
\footnote{To train \methodname{} for a single subject we use $\approx0.006\%$ of the dataset used by AVA~\cite{cao2022authentic}.}. 
We use Peak Signal To Noise Ratio (PSNR)~\cite{avcibas2002statistical}, Structural Similarity Index (SSIM)~\cite{wang2003multiscale} and Learned Perceptual Image Patch Similarity (LPIPS)~\cite{zhang2018perceptual} to measure the performance of the models. Each of the rendered images has a resolution of $334{\times}512$ pixels.\looseness=-1

As baselines, we use the following approaches: the original, static NeRF~\cite{mildenhall2020nerf}, NeRF conditioned on an expression code concatenated with input points~$\pos$, NeRFies~\cite{park2021nerfies}, HyperNeRF\footnote{We use two architectures proposed by Park~\etal~\cite{park2021hypernerf}.}~\cite{park2021hypernerf}, and VolTeMorph~\cite{garbin2022voltemorph}.  We replace the learnable code in NeRFies and HyperNeRF with the expression code $\expression$ from the parametric model. Since VolTeMorph can be trained on multiple frames, which should lead to averaging of the output colors, we split it into two regimes: one trained on the most extreme expression\footnote{We manually select one frame that has the most visible wrinkles.} (VolTeMorph$_1$) and the another trained on all available expressions (VolTeMorph$_\text{avg}$)\footnote{We do not compare to NeRFace~\cite{gafni2021dynamic} and NHA~\cite{neural_head_avatars} as VolTeMorph~\cite{garbin2022voltemorph} performs better quantitatively than these methods.}. We use both of these baselines as VolTeMorph was originally designed for a single-frame scenario. By using two versions, we show that it is not trivial to extend it to multiple expressions.%

\subsection{Realistic Human Captures}
\label{subsec:realistic-human-captures}
\noindent\textbf{Novel expression synthesis.}
We extract eight multi-view frames from the Multiface dataset~\cite{wuu2022multiface}, each of a different expression. Five of these expressions serve as training data, and the rest are used for evaluation. After training, we can extrapolate from the training expressions by modifying the expression vector~$\expression$. 
We use the remaining three expressions: moving mouth left and right, and puffing cheeks, to evaluate the capability of the models to reconstruct other expressions. 
In~\cref{fig:qualitative-comparison} we show that \methodname{} is the only method capable of rendering convincing wrinkles dynamically, depending on the input expression. 
\methodname{} performs favorably compared to the baselines~(see \cref{tab:quantitative-results}).

\noindent\textbf{Casual expressions.} The Multiface dataset contains sequences where the subject follows a script of expressions to show during the capture. Each of these captures contains between 1000 and 2000 frames. This experiment tests whether a model can interpolate between the training expressions smoothly and generalize beyond the training data. Quantitative results are shown in \cref{tab:quantitative-results}. Our approach performs best all the settings.  
See animations in the \supplementary{} for a comparison of rendered frames across all methods.

\subsection{Modeling Objects Beyond Faces}
We show that our method can be applied beyond face modeling. We prepare two datasets containing 96 views per frame of bending and twisting cylinders made of a rubber-like material (24 and 72 temporal frames, respectively). When bent or twisted, the cylinders reveal pose-dependent details. The expression vector $\expression$ now encodes time: 0 if the cylinder is in the canonical pose, 1 if it is posed, and any values between $[0, 1]$ for the transitioning stage. We select expressions $\{0, 0.5, 1.0\}$ as a training set (for VolTeMorph$_1$ we use $1.0$ only). For evaluation, we take every fourth frame from the full sequence using cameras from the bottom and both sides of the object.
We take the mesh directly from Houdini~\cite{xu2014houdini}, which we use for wrinkle simulation, and render the images in Blender~\cite{blender2022}.
We show quantitative results in~\cref{tab:quantitative-results} for the bending cylinder, and a comparison of the inferred images in~\cref{fig:synthetic-qualitative} for the twisted one\footnote{Our motivation is that it is easier to show pose-dependent deformations on twisting as it affects the object globally, while the bending cannot be modeled by all the baselines due to the non-stationary effects.}. \methodname{} accurately captures the transition from the rest configuration to the deformed state of the cylinder, rendering high-frequency details where required. All other approaches struggle with interpolation between states. VolTeMorph$_1$ (trained on a single extreme pose) renders wrinkles even when the cylinder is not twisted.

\subsection{Ablations}
We check how the neighborhood size~$|\neighbourhood(\vertex)|$ and the application of the smoothing influence the performance of our method. We show the results in~\cref{tab:ablation-study}. \methodname{} works best in most cases when considering a relatively wide neighborhood for the tetrahedral features\footnote{Larger neighborhood sizes caused out-of-memory errors on our NVIDIA 2080Ti GPU.}.
Laplacian smoothing consistently improves the quality across all the datasets~(see~\cref{fig:laplacian-smoothing}). We additionally present in the \supplementary{} how the number of expressions used for training affects the results.

\subsection{Failure Cases}
\label{subsec:failures}
\newcommand{\leftimageheight}{4.5cm}
\newcommand{\rightimageheight}{4.5cm}

\renewcommand{\versionone}{
    \begin{tikzpicture}[
        >=stealth',
        overlay/.style={
          anchor=south west, 
          draw=black,
          rectangle, 
          line width=0.8pt,
          outer sep=0,
          inner sep=0,
        },
    ]
        \matrix[
        matrix of nodes, 
        column sep=2pt, 
        row sep=0pt, 
        ampersand replacement=\&, 
        inner sep=0, 
        outer sep=0,
        ] (main) {
            \includegraphics[height=\leftimageheight]{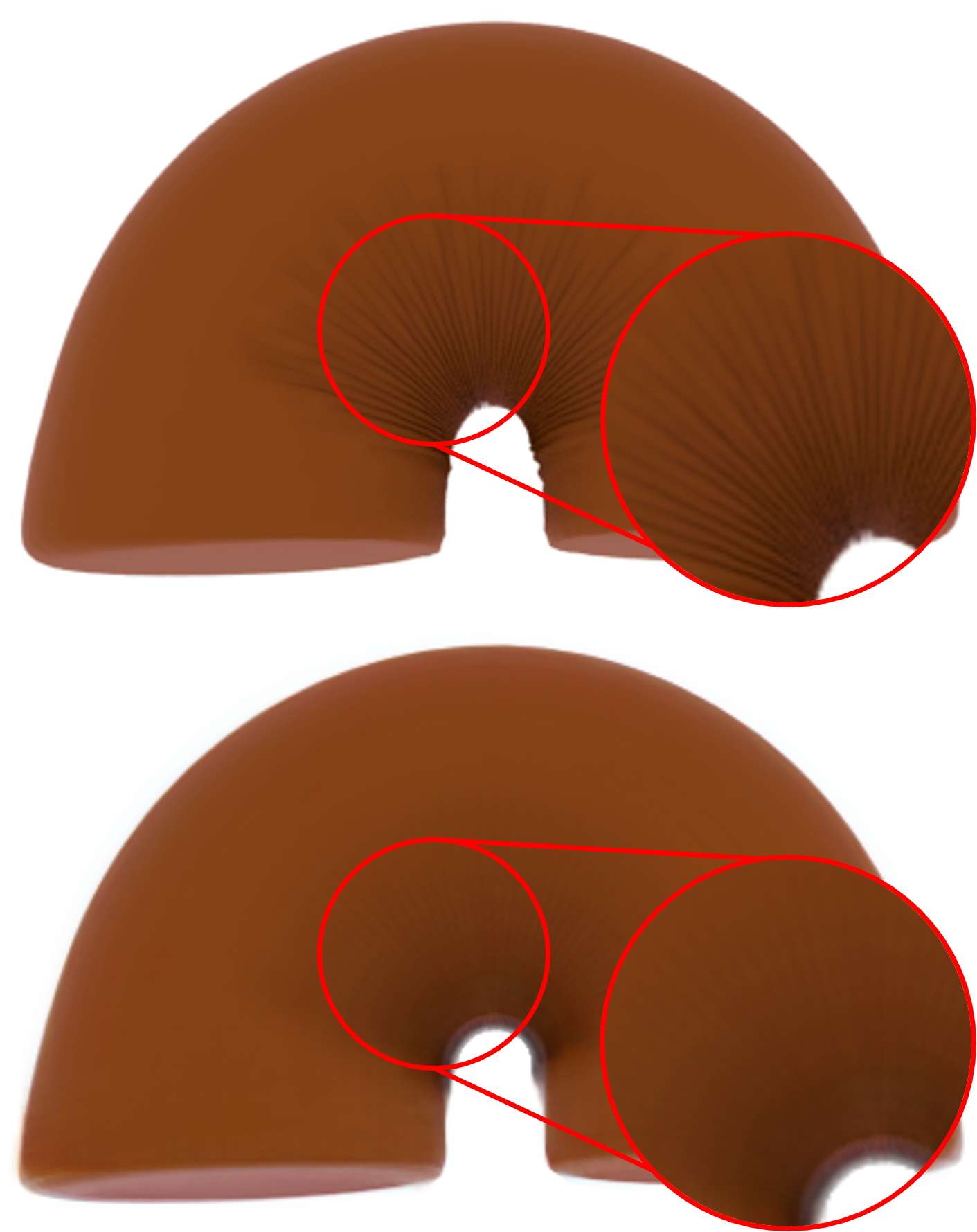} \&
            \includegraphics[height=\rightimageheight]{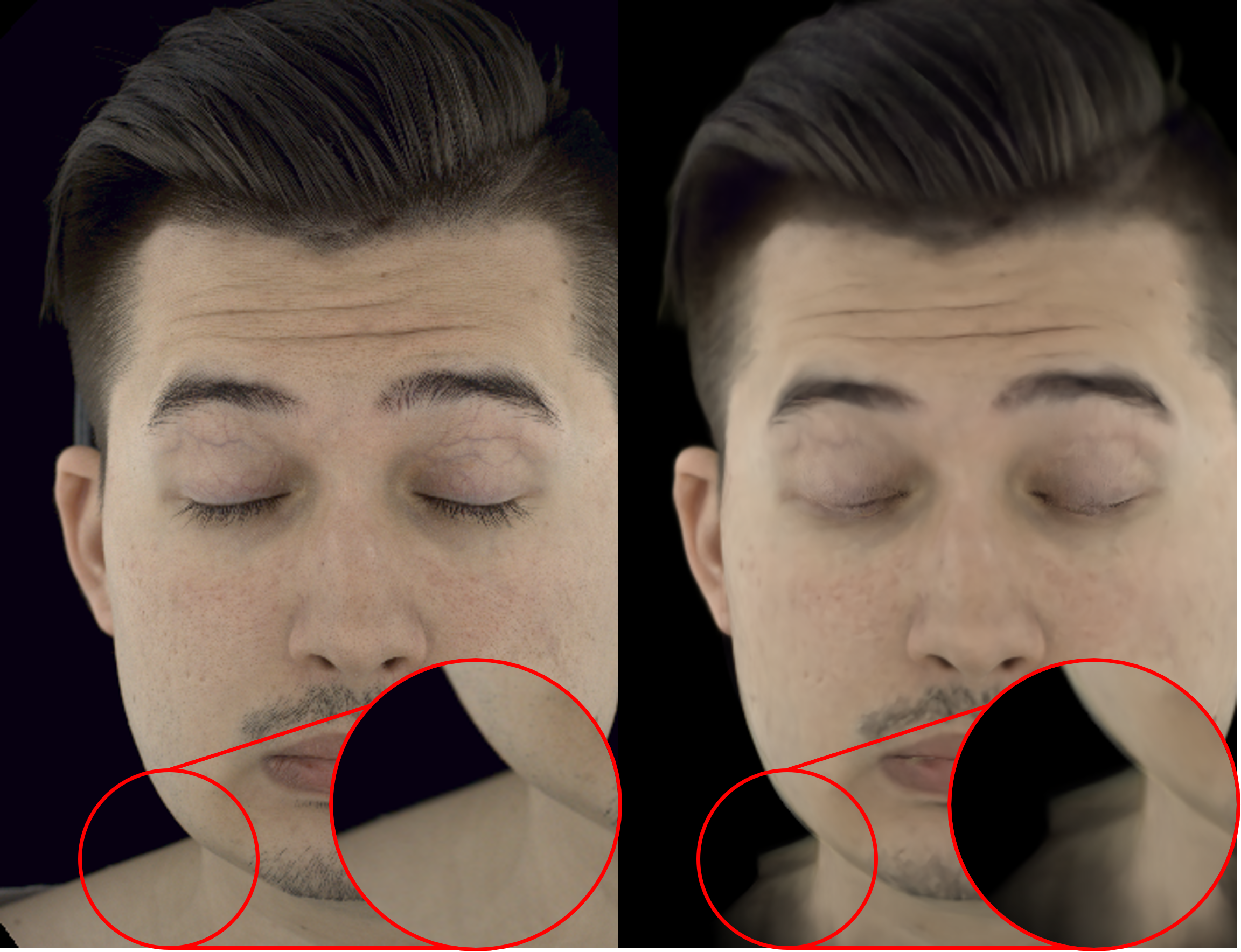} \\
        };
        \node[above=0.0em of main-1-1.north, align=center, anchor=south]{Low contrast};
        \node[above=0.0em of main-1-2.north, align=center, anchor=south]{Inaccurate off-the-shelf tracker};
    \end{tikzpicture}
}
\begin{figure}[t]
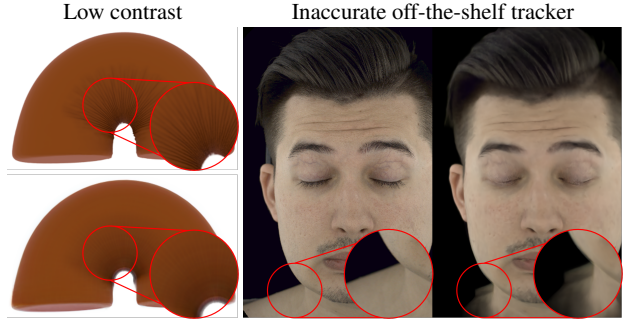

    \centering
    \resizebox{\linewidth}{!}{\versionone}
    \caption{\textbf{Failure cases} -- {%
    We show failure cases for our proposed approach. \textit{Left:} In the presence of wrinkles in low-contrast images, \methodname{} takes longer to converge to make wrinkles visible. We show the ground truth on the top, and rendering after training $7{\times}10^5$ steps on the bottom. In contrast, we rendered images in~\Cref{fig:synthetic-qualitative} after $2{\times}10^5$ steps.  \textit{Right:} \methodname{} inherits issues from VolTeMorph~\cite{garbin2022voltemorph}, which relies on the initial fit of the face mesh. If the fit is inaccurate, artifacts appear in the final render.
    }}
    \label{fig:failure-cases}
    \vspace{-1.25em}
\end{figure}
While \methodname{} offers significant advantages for rendering realistic and dynamic high-frequency details, it falls short in some scenarios (see~\cref{fig:failure-cases}). 
One of the issues arises when the contrast between wrinkles and the subject's skin color is low. In those instances, we observe a much longer time to convergence. 
Moreover, as we build \methodname{} on VolTeMorph, we also inherit some of its problems. Namely, the method heavily relies on the initial fit of the parametric model -- any inaccuracy leads to ghosting artifacts or details on the face that jump between frames. 

\section{Conclusions}
\label{ref:conclusion}
We present a general approach, \methodname{}, for rendering high-frequency expression-dependent details using NeRFs. 
\methodname{} draws inspiration from classical computer graphics by blending expressions from the training data to render expressions unseen during training. We show that \methodname{} renders images in a controllable and interpretable manner for novel expressions and can be applied to render human avatars learned from publicly available datasets. 
We additionally discuss the potential misuse of our work in the \supplementary{}.

\iftoggle{cvprfinal}{%
\section{Acknowledgements}
\label{sec:acknowledgements}
The work was partly supported by the National Sciences and Engineering Research Council of Canada (NSERC), the Digital Research Alliance of Canada, and Microsoft Mesh Labs. This research was funded by Microsoft Research through the EMEA PhD Scholarship Programme. We thank NVIDIA Corporation for granting us access to GPUs through NVIDIA's Academic Hardware Grants Program. This research was partially funded by National Science Centre, Poland (grant no 2020/39/B/ST6/01511 and 2022/45/B/ST6/02817).
}{}

{\small
\bibliographystyle{ieee_fullname}
\bibliography{macros,main}
}

\appendix

\setcounter{page}{1}

\twocolumn[
\centering
\Large
\textbf{\methodname{}: Few-Shot Example-Driven Facial Modeling} \\
\vspace{0.5em}Supplementary Material \\
\vspace{1.0em}
] %
\appendix

\section{Potential social impact}
Our motivation for this work was to enable the creation of 3D avatars that could be used as communication devices in the remote working era. As our approach stems from blendshapes~\cite{lewis2014practice}, these avatars are easily adjustable via texture coloring and may be used for entertainment. 
We note, however, that the potential misuse of our work includes using it as deep fakes. We highly discourage such usage. One of our future directions includes detecting fake images generated by our method. 
At the same time, we highlight the importance of \methodname{}---in the presence of closed technologies~\cite{ma2021pixel,cao2022authentic}, it is crucial to democratize techniques for personalized avatar creation. We achieve that by limiting the required data volume to train a single model. As history shows, when given an open, readily available technology for generative modeling of images~\cite{rombach2022high}, users can scrutinize it with unprecedented thoroughness, thus raising the general awareness of potential misuses. 

\section{Concurrent Works}
Gao~\etal~\cite{gao2022reconstructing} and Xu~\etal~\cite{xu2022manvatar} also use an interpolation between known expressions to combine multiple neural radiance fields trained for those expressions. However, their approach interpolates between grids of latent vectors~\cite{mueller2022instant} globally. The interpolation weights are taken from blendshape coefficients. 

Zielonka~\etal~\cite{zielonka2022instant} use a parametric head model to canonicalize 3D points similarly to our ends. However, instead of building a tetrahedral cage around the head, they smoothly assign each face triangle to 3D points. Then they canonicalize points using transformations that each of the assigned triangles undergoes for a given expression. They concatenate 3D points with the expression code from FLAME~\cite{li2017flame} to model expression-dependent effects.

\begin{table}[!t]
    \centering
    \resizebox{\linewidth}{!}{
        \begin{tabular}{ccccccc}
        \toprule
        \multirow{2}[2]{*}{\# expr.} & \multicolumn{3}{c}{Casual Expressions} & \multicolumn{3}{c}{Novel Pose Synthesis} \\
        \cmidrule{2-7}
         &  PSNR $\uparrow$ & SSIM $\uparrow$ & LPIPS $\downarrow$  &  PSNR $\uparrow$ & SSIM $\uparrow$ & LPIPS $\downarrow$ \\
        \midrule
        $\nExpr{=}1$&                              27.5834 &                             0.9028 &                             0.0834 &                            28.7589 &                             0.9147 &                             0.0806  \\
        $\nExpr{=}2$&                              27.6783 &                             0.9026 &                             0.0856 &                            29.2859 &                             0.9186 &                             0.0803  \\
        $\nExpr{=}3$&                              27.9137 &                             0.9054 &                             0.0819 &                            29.8551 &                             0.9279 &                             0.0728  \\
        $\nExpr{=}4$&                              27.8140 &                             0.9055 &                             0.0815 & \cellcolor{secondbestcolor}30.1543 &  \cellcolor{secondbestcolor}0.9336 &  \cellcolor{secondbestcolor}0.0701  \\
        $\nExpr{=}5$&                              28.0254 &                             0.9110 &   \cellcolor{firstbestcolor}0.0778 &  \cellcolor{firstbestcolor}30.4721 &   \cellcolor{firstbestcolor}0.9372 &   \cellcolor{firstbestcolor}0.0688  \\
        \midrule
        $\nExpr{=}6$&                              28.0517 &                             0.9091 &  \cellcolor{secondbestcolor}0.0813 & -- & -- & -- \\
        $\nExpr{=}7$&   \cellcolor{secondbestcolor}28.2004 &  \cellcolor{secondbestcolor}0.9115 &                             0.0823 & -- & -- & -- \\
        $\nExpr{=}8$&    \cellcolor{firstbestcolor}28.2542 &   \cellcolor{firstbestcolor}0.9124 &                             0.0830 & -- & -- & -- \\
        \bottomrule
        \end{tabular}
    }
    \caption{\textbf{Number of training expressions} -- {%
            We ablate over the number of training expressions. We evaluate the model on the captures from the Multiface dataset~\cite{wuu2022multiface}. We run the model for each possible expression combination for a given $\nExpr$ and average the results. The best results are colored in \mycoloredbox{firstbestcolor} and the second best in \mycoloredbox{secondbestcolor}. Increasing the number of available training expressions consistently improves the results. However, using $\nExpr{=}5$ expressions saturates the quality and using $\nExpr{>}5$ brings diminishing improvements.  We do not report ``Novel Pose Synthesis'' for $\nExpr{>}5$ as we use validation expressions and poses to train those models (refer to~\cref{subsec:realistic-human-captures} for more details). %
        }
    }
    \vspace{-1.2em}
    \label{tab:ablation-num-expressions-fix}
    
\end{table}

\section{Additional results}
\subsection{Ablating number of expressions}
We ablate over the number of used expressions during the training. To evaluate the effect of the number of expressions, we add consecutive frames to the training set (starting from a single, neutral one), \ie, the training set has $\iExpr{<}\nExpr$ expressions. We train \methodname{} for such a set for each subject separately. We then average the results for a given $\iExpr$ across subjects. We present the results in~\cref{tab:ablation-num-expressions-fix}. When selecting the training expressions, we aim to choose those that show all wrinkles when combined. We can see from~\cref{fig:qualitative-ablation-expressions} that if removed, \eg, the expressions with eyebrows raised, then the model cannot render wrinkles on the forehead. In summary, increasing the number of expressions improves the quality results with diminishing returns when $\nExpr{>}5$, while $\nExpr{=}5$ provides a sufficient trade-off between the data capture cost and the quality.%

\begin{figure}[t]
    \centering
    \includegraphics[width=\linewidth,trim={0 2em 0 2em}, clip]{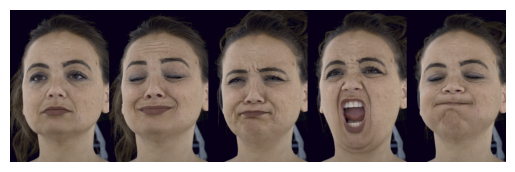}
    \caption{\textbf{Training frames} -- In~\cref{ref:experiments}, we show results for the \methodname{} trained on $\nExpr{=}5$ expressions. The images represent these expressions for one of the subjects. For each subject, we selected similar expressions to show all possible wrinkles when combined. Please note that we also include a ``neutral'' expression (the first from the left)---it is necessary to enable 
    the learning of
    a face without any wrinkles.}
    \label{fig:supplementary-training-frames}
\end{figure}
\subsection{Training frames}
We present in~\cref{fig:supplementary-training-frames} example training frames for one of the subjects. Each frame is a multi-view frame captured with ${\approx} 35$ cameras (the number of available cameras varied slightly between subjects).

\begin{table}[!t]
    \centering
    \resizebox{\linewidth}{!}{
        \begin{tabular}{lcccccc}
        \toprule
        \multirow{2}[2]{*}{Method} & \multicolumn{3}{c}{Casual Expressions} & \multicolumn{3}{c}{Novel Pose Synthesis} \\
        \cmidrule(lr){2-4}\cmidrule(lr){5-7}
        & PSNR $\uparrow$ & SSIM $\uparrow$ & LPIPS $\downarrow$ & PSNR $\uparrow$ & SSIM $\uparrow$ & LPIPS $\downarrow$ \\
        \midrule
NeRF \cite{mildenhall2020nerf}                     &                              22.0060 &                             0.6556 &                             0.3222&                             23.8077 &                             0.7448 &                             0.2779   \\
Conditioned NeRF \cite{mildenhall2020nerf}         &                              21.0846 &                             0.6280 &                             0.3042&                             22.9991 &                             0.7261 &                             0.2362   \\
NeRFies \cite{park2021nerfies}                     &                              20.7004 &                             0.6076 &                             0.3579&                             23.0123 &                             0.7253 &                             0.2840   \\
HyperNeRF-AP \cite{park2021hypernerf}              &                              20.8105 &                             0.6214 &                             0.3504&                             22.8193 &                             0.7185 &                             0.2689   \\
HyperNeRF-DS \cite{park2021hypernerf}              &                              20.8847 &                             0.6111 &                             0.3656&                             23.0075 &                             0.7259 &                             0.2729   \\
\midrule
VolTeMorph$_1$ \cite{garbin2022voltemorph}         &                              21.3265 &                             0.7091 &                             0.2706&                             22.3007 &                             0.7795 &  \cellcolor{secondbestcolor}0.2281   \\
VolTeMorph$_\text{avg}$\cite{garbin2022voltemorph} &   \cellcolor{secondbestcolor}22.0759 &  \cellcolor{secondbestcolor}0.7755 &  \cellcolor{secondbestcolor}0.2615&  \cellcolor{secondbestcolor}23.8974 &  \cellcolor{secondbestcolor}0.8458 &                             0.2302   \\
\midrule
\textbf{\methodname{}}                             &    \cellcolor{firstbestcolor}22.8982 &   \cellcolor{firstbestcolor}0.7954 &   \cellcolor{firstbestcolor}0.2256&   \cellcolor{firstbestcolor}24.4432 &   \cellcolor{firstbestcolor}0.8477 &   \cellcolor{firstbestcolor}0.2052   \\
\bottomrule  
        \end{tabular}
    }
    \caption{\textbf{Quantitative results without masking} -- 
    Similarly to \cref{tab:quantitative-results}, we compare \methodname{} to other related approaches. However, we calculate the results over the whole image space, without removing the background. \methodname{} and VolTeMorph~\cite{garbin2022voltemorph} model the background as a separate NeRF-based~\cite{mildenhall2020nerf} network. The points that do not fall into the tetrahedral mesh are assigned to the background. As the network overfits to sparse training views, it poorly extrapolates to novel expressions (as the new head pose or expression may reveal some unknown parts of the background) and views. At the same time, all other baselines do not have any mechanism to disambiguate the background and the foreground.
    }
    \label{tab:quantitative-results-without-masking}
\end{table}

\subsection{Quantitative results with background}
We compare \methodname{} and the baselines similarly to~\cref{subsec:realistic-human-captures}. However, in this experiment, we deliberately include the background in metric calculation. We show the results in \cref{tab:quantitative-results-without-masking}. In all the cases, \methodname{} performs best even though the method was not designed to model the background accurately. Additionally, as HyperNeRF~\cite{park2021hypernerf}, NeRFies~\cite{park2021nerfies}, and NeRF~\cite{mildenhall2020nerf} do not have any mechanism to disambiguate between the foreground and the background, the metrics are significantly worse when including the latter.

\subsection{Additional qualitative results}

We show in~\cref{fig:qualitative-other-baselines} results of baselines that do not rely on parametric models of the face~\cite{li2017flame}. Compared to \methodname{}, they cannot render high-fidelity faces. The issue comes from the assumed data sparsity---those approaches rely on the interpolation in the training data. As we assume access to just a few frames, there is no continuity in the training data that would guide them to interpolate between known expressions. \methodname{} presents superior results given novel expressions even with such a sparse dataset.
See the attached video and \texttt{index.html} file for more qualitative results.

\clearpage

\renewcommand{\imagewidth}{3.3cm}
\renewcommand{\imageheight}{4.0263671874999996cm}
\renewcommand{\smallimagewidth}{0.8cm}
\renewcommand{\expressionsep}{0em}

\newcommand{\firstindex}{0189}
\newcommand{\secondindex}{0091}
\newcommand{\thirdindex}{0505}
\newcommand{\fourthindex}{0852}

\newcommand{\supimageoneindex}[2]{%
    \includegraphics[width=\imagewidth, trim={0 0 0 0}, clip]{assets/supplementary/expressions/#1_2183941_selected_rom_#2_\firstindex.png}%
}
\newcommand{\supimagetwoindex}[2]{%
    \includegraphics[width=\imagewidth, trim={0 0 0 0}, clip]{assets/supplementary/expressions/#1_5372021_selected_rom_#2_\secondindex.png}%
}
\newcommand{\supimagethreeindex}[2]{%
    \includegraphics[width=\imagewidth, trim={0 0 0 0}, clip]{assets/supplementary/expressions/#1_6795937_selected_rom_#2_\thirdindex.png}%
}
\newcommand{\supimagefourindex}[2]{%
    \includegraphics[width=\imagewidth, trim={0 0 0 0}, clip]{assets/supplementary/expressions/#1_7889059_selected_rom_#2_\fourthindex.png}%
}

\renewcommand{\versionone}{
    \begin{tikzpicture}[
        >=stealth',
        overlay/.style={
          anchor=south west, 
          draw=black,
          rectangle, 
          line width=0.8pt,
          outer sep=0,
          inner sep=0,
        },
    ]
        \matrix[
        matrix of nodes, 
        column sep=0pt, 
        row sep=0pt, 
        ampersand replacement=\&, 
        inner sep=0, 
        outer sep=0,
        draw=black,
        rectangle,
        line width=0.5pt
        ] (gt) {
            \supimageoneindex{gt}{0} \\
            \supimagetwoindex{gt}{0} \\
            \supimagethreeindex{gt}{0} \\
            \supimagefourindex{gt}{0} \\
        };
        
        \matrix[
        matrix of nodes, 
        column sep=0pt, 
        row sep=0pt, 
        ampersand replacement=\&, 
        inner sep=0, 
        outer sep=0,
        right=0pt of gt-1-1.east,
        draw=black,
        rectangle,
        line width=0.5pt
        ] (k1) {
            \supimageoneindex{bv_aux}{0} \&
            \supimageoneindex{bv_aux}{0-1} \&
            \supimageoneindex{bv_aux}{0-1-2} \&
            \supimageoneindex{bv_aux}{0-1-2-3} \&
            \supimageoneindex{bv_aux}{0-1-2-3-4} \\
        };
        
        \matrix[
        matrix of nodes, 
        column sep=0pt, 
        row sep=0pt, 
        ampersand replacement=\&, 
        inner sep=0, 
        outer sep=0,
        right=0pt of gt-2-1.east,
        draw=black,
        rectangle,
        line width=0.5pt
        ] (k2) {
            \supimagetwoindex{bv_aux}{0} \&
            \supimagetwoindex{bv_aux}{0-1} \&
            \supimagetwoindex{bv_aux}{0-1-2} \&
            \supimagetwoindex{bv_aux}{0-1-2-3} \&
            \supimagetwoindex{bv_aux}{0-1-2-3-4} \\
        };
        
        \matrix[
        matrix of nodes, 
        column sep=0pt, 
        row sep=0pt, 
        ampersand replacement=\&, 
        inner sep=0, 
        outer sep=0,
        right=0pt of gt-3-1.east,
        draw=black,
        rectangle,
        line width=0.5pt
        ] (k3) {
            \supimagethreeindex{bv_aux}{0} \&
            \supimagethreeindex{bv_aux}{0-1} \&
            \supimagethreeindex{bv_aux}{0-1-2} \&
            \supimagethreeindex{bv_aux}{0-1-2-3} \&
            \supimagethreeindex{bv_aux}{0-1-2-3-4} \\
        };
        
        \matrix[
        matrix of nodes, 
        column sep=0pt, 
        row sep=0pt, 
        ampersand replacement=\&, 
        inner sep=0, 
        outer sep=0,
        right=0pt of gt-4-1.east,
        draw=black,
        rectangle,
        line width=0.5pt
        ] (k4) {
            \supimagefourindex{bv_aux}{0} \&
            \supimagefourindex{bv_aux}{0-1} \&
            \supimagefourindex{bv_aux}{0-1-2} \&
            \supimagefourindex{bv_aux}{0-1-2-3} \&
            \supimagefourindex{bv_aux}{0-1-2-3-4} \\
        };
        
        \node[above=0em of gt-1-1.north, align=center, anchor=south]{Ground Truth};
        \node[above=0em of k1-1-1.north, align=center, anchor=south]{$K={1}$};
        \node[above=0em of k1-1-2.north, align=center, anchor=south]{$K={2}$};
        \node[above=0em of k1-1-3.north, align=center, anchor=south]{$K={3}$};
        \node[above=0em of k1-1-4.north, align=center, anchor=south]{$K={4}$};
        \node[above=0em of k1-1-5.north, align=center, anchor=south]{$K={5}$};
        
        \node[left=0em of gt-1-1.west, align=center, anchor=east]{\rotatebox{90}{\textsc{Subject} 2183941}};
        \node[above=0em of gt-2-1.west, align=center, anchor=east]{\rotatebox{90}{\textsc{Subject} 5372021}};
        \node[above=0em of gt-3-1.west, align=center, anchor=east]{\rotatebox{90}{\textsc{Subject} 6795937}};
        \node[above=0em of gt-4-1.west, align=center, anchor=east]{\rotatebox{90}{\textsc{Subject} 7889059}};
    \end{tikzpicture}
}

\begin{figure*}[htb]
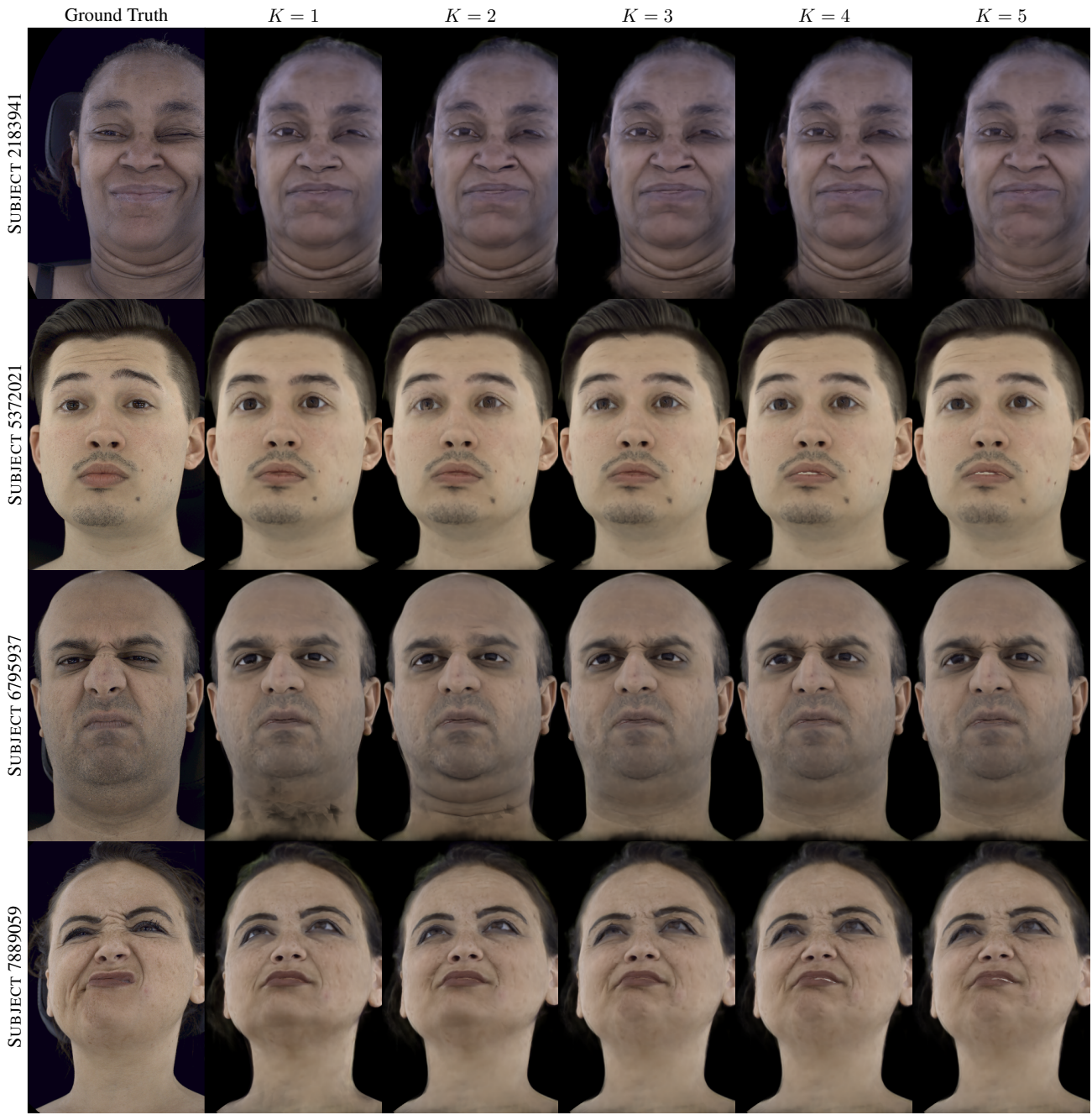

    \centering
    \resizebox{\linewidth}{!}{
        \versionone
    }
    \caption{\textbf{Qualitative ablation over the number of training expressions} -- {%
        We show qualitatively how the number of training expressions $\nExpr$ affects the rendering quality. The first row shows the ground truth images. All other consecutive rows show the images rendered with \methodname{} while  increasing the number of training expressions. The last row, $\nExpr{=}5$ corresponds to the results presented in the main part of the article. The subject's naming follows the convention introduced in the Multiface repository~\cite{wuu2022multiface}. Please refer to~\cref{tab:ablation-num-expressions-fix} for quantitative results.
    }}
    \vspace{-1.2em}
    \label{fig:qualitative-ablation-expressions}
\end{figure*}

\renewcommand{\imagewidth}{2.7cm}
\renewcommand{\imageheight}{4.0263671874999996cm}
\renewcommand{\smallimagewidth}{0.8cm}
\renewcommand{\expressionsep}{0em}

\renewcommand{\firstindex}{0580}
\renewcommand{\secondindex}{0378}
\renewcommand{\thirdindex}{0226}
\renewcommand{\fourthindex}{0720}

\renewcommand{\supimageoneindex}[1]{%
    \includegraphics[width=\imagewidth, trim={0 0 0 0}, clip]{assets/supplementary/baselines/#1_2183941_selected_rom_\firstindex.png}%
}
\renewcommand{\supimagetwoindex}[1]{%
    \includegraphics[width=\imagewidth, trim={0 0 0 0}, clip]{assets/supplementary/baselines/#1_5372021_selected_rom_\secondindex.png}%
}
\renewcommand{\supimagethreeindex}[1]{%
    \includegraphics[width=\imagewidth, trim={0 0 0 0}, clip]{assets/supplementary/baselines/#1_6795937_selected_rom_\thirdindex.png}%
}
\renewcommand{\supimagefourindex}[1]{%
    \includegraphics[width=\imagewidth, trim={0 0 0 0}, clip]{assets/supplementary/baselines/#1_7889059_selected_rom_\fourthindex.png}%
}

\renewcommand{\versionone}{
    \begin{tikzpicture}[
        >=stealth',
        overlay/.style={
          anchor=south west, 
          draw=black,
          rectangle, 
          line width=0.8pt,
          outer sep=0,
          inner sep=0,
        },
    ]
        \matrix[
        matrix of nodes, 
        column sep=0pt, 
        row sep=0pt, 
        ampersand replacement=\&, 
        inner sep=0, 
        outer sep=0,
        draw=black,
        rectangle,
        line width=0.5pt
        ] (gt) {
            \supimageoneindex{gt} \\
            \supimagetwoindex{gt} \\
            \supimagethreeindex{gt} \\
            \supimagefourindex{gt} \\
        };
        
        \matrix[
        matrix of nodes, 
        column sep=0pt, 
        row sep=0pt, 
        ampersand replacement=\&, 
        inner sep=0, 
        outer sep=0,
        right=0pt of gt-1-1.east,
        draw=black,
        rectangle,
        line width=0.5pt
        ] (k1) {
            \supimageoneindex{bv_aux} \&
            \supimageoneindex{nerf}   \&
            \supimageoneindex{dnerf} \&
            \supimageoneindex{nerfies} \&
            \supimageoneindex{hypernerf_static}    \&
            \supimageoneindex{hypernerf_dynamic} \\
        };
        
        \matrix[
        matrix of nodes, 
        column sep=0pt, 
        row sep=0pt, 
        ampersand replacement=\&, 
        inner sep=0, 
        outer sep=0,
        right=0pt of gt-2-1.east,
        draw=black,
        rectangle,
        line width=0.5pt
        ] (k2) {
            \supimagetwoindex{bv_aux} \&
            \supimagetwoindex{nerf}   \&
            \supimagetwoindex{dnerf} \&
            \supimagetwoindex{nerfies} \&
            \supimagetwoindex{hypernerf_static}    \&
            \supimagetwoindex{hypernerf_dynamic} \\
        };
        
        \matrix[
        matrix of nodes, 
        column sep=0pt, 
        row sep=0pt, 
        ampersand replacement=\&, 
        inner sep=0, 
        outer sep=0,
        right=0pt of gt-3-1.east,
        draw=black,
        rectangle,
        line width=0.5pt
        ] (k3) {
            \supimagethreeindex{bv_aux} \&
            \supimagethreeindex{nerf}   \&
            \supimagethreeindex{dnerf} \&
            \supimagethreeindex{nerfies} \&
            \supimagethreeindex{hypernerf_static}    \&
            \supimagethreeindex{hypernerf_dynamic} \\
        };
        
        \matrix[
        matrix of nodes, 
        column sep=0pt, 
        row sep=0pt, 
        ampersand replacement=\&, 
        inner sep=0, 
        outer sep=0,
        right=0pt of gt-4-1.east,
        draw=black,
        rectangle,
        line width=0.5pt
        ] (k4) {
            \supimagefourindex{bv_aux} \&
            \supimagefourindex{nerf}   \&
            \supimagefourindex{dnerf} \&
            \supimagefourindex{nerfies} \&
            \supimagefourindex{hypernerf_static}    \&
            \supimagefourindex{hypernerf_dynamic} \\
        };
        
        \node[above=0em of gt-1-1.north, align=center, anchor=south]{Ground Truth};
        \node[above=0em of k1-1-1.north, align=center, anchor=south]{\textbf{\methodname{}}};
        \node[above=0em of k1-1-2.north, align=center, anchor=south]{NeRF};
        \node[above=0em of k1-1-3.north, align=center, anchor=south]{NeRF+expr};
        \node[above=0em of k1-1-4.north, align=center, anchor=south]{NeRFies};
        \node[above=0em of k1-1-5.north, align=center, anchor=south]{HyperNeRF-AP};
        \node[above=0em of k1-1-6.north, align=center, anchor=south]{HyperNeRF-DS};
        
        \node[left=0em of gt-1-1.west, align=center, anchor=east]{\rotatebox{90}{\textsc{Subject} 2183941}};
        \node[above=0em of gt-2-1.west, align=center, anchor=east]{\rotatebox{90}{\textsc{Subject} 5372021}};
        \node[above=0em of gt-3-1.west, align=center, anchor=east]{\rotatebox{90}{\textsc{Subject} 6795937}};
        \node[above=0em of gt-4-1.west, align=center, anchor=east]{\rotatebox{90}{\textsc{Subject} 7889059}};
    \end{tikzpicture}
}

\begin{figure*}[htb]
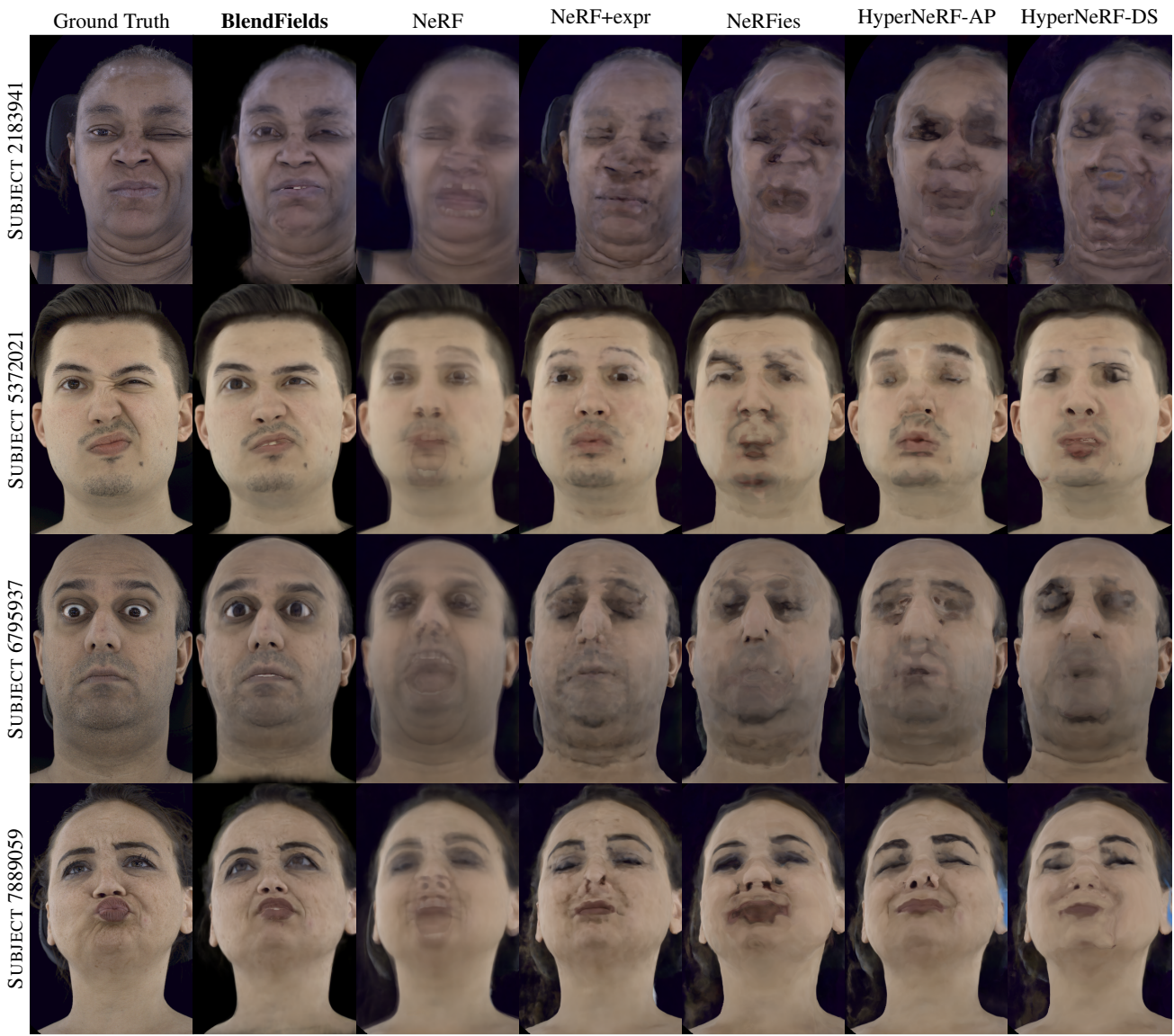

    \centering
    \resizebox{\linewidth}{!}{\versionone}
    \caption{\textbf{Comparison to strictly data-driven approaches} -- 
        We compare \methodname{} to other baselines that do not rely on mesh-driven rendering: NeRF~\cite{mildenhall2020nerf}, NeRF conditioned on the expression code (NeRF+expr)~\cite{mildenhall2020nerf}, NeRFies~\cite{park2021nerfies}, and HyperNeRF-AP/DS~\cite{park2021hypernerf}. As a static model, NeRF converges to an average face from available ($\nExpr{=}5$) expressions. All other baselines exhibit severe artifacts compared to \methodname{}. Those baselines rely on the data continuity in the training set (\eg, from a video), and cannot generalize to any other expression. Please see the supplemented video for the animations.
    }
    \vspace{-1.2em}
    \label{fig:qualitative-other-baselines}
\end{figure*}

\end{document}